\documentclass{article}
\pdfoutput=1

\PassOptionsToPackage{numbers}{natbib}
\usepackage[preprint]{neurips_2023}

\usepackage[utf8]{inputenc} 
\usepackage[T1]{fontenc}    
\usepackage{hyperref}       
\usepackage{url}            
\usepackage{booktabs}       
\usepackage{amsfonts}       
\usepackage{nicefrac}       
\usepackage{microtype}      
\usepackage{xcolor}         

\usepackage[export]{adjustbox}
\usepackage{amsmath}
\usepackage{amsthm}
\usepackage{bm}
\usepackage{graphicx}
\usepackage{mathtools}
\usepackage{multirow}
\usepackage{nicefrac}       
\usepackage{pifont}
\usepackage{pgfplots}
\usepackage{tikz}
\usepackage{subcaption}
\usepackage{wrapfig}
\pgfplotsset{
    compat=1.17
}

\theoremstyle{plain}

\theoremstyle{definition}

\theoremstyle{remark}

\newcommand\ourmethod[0]{\textsc{D\&D}}

\newcommand{\cutsectiondown}{\vspace*{-0.12in}}
\newcommand{\cutsubsectionup}{\vspace*{-0.1in}}
\newcommand{\cutsubsectiondown}{\vspace*{-0.07in}}

\newcommand{\BI}{\bm{I}}

\newcommand{\BR}{\bm{R}}

\newcommand{\BZ}{\bm{Z}}

\newcommand{\Bepsilon}{\bm{\epsilon}}

\newcommand{\loss}{\mathcal{L}}

\newcommand{\R}{\mathbb{R}}

\newcommand{\E}{\mathbb{E}}

\title{3D Denoisers are Good 2D Teachers: Molecular Pretraining via Denoising and Cross-Modal Distillation}

\author{%
  Sungjun Cho$^1$ \quad Dae-Woong Jeong$^1$ \quad Sung Moon Ko$^1$ \quad Jinwoo Kim$^2$\\
  \textbf{Sehui Han}$^{1}$ \quad \textbf{Seunghoon Hong}$^2$ \quad \textbf{Honglak Lee}$^1$ \quad \textbf{Moontae Lee}$^{1,3}$\\
  $^1$LG AI Research \quad $^2$KAIST \quad $^3$University of Illinois Chicago\\
}

\begin{document}

\maketitle

\begin{abstract}
Pretraining molecular representations from large unlabeled data is essential for molecular property prediction due to the high cost of obtaining ground-truth labels. While there exist various 2D graph-based molecular pretraining approaches, these methods struggle to show statistically significant gains in predictive performance. Recent work have thus instead proposed 3D conformer-based pretraining under the task of denoising, which led to promising results. During downstream finetuning, however, models trained with 3D conformers require accurate atom-coordinates of previously unseen molecules, which are computationally expensive to acquire at scale. In light of this limitation, we propose \ourmethod{}, a self-supervised molecular representation learning framework that pretrains a 2D graph encoder by distilling representations from a 3D denoiser. With denoising followed by cross-modal knowledge distillation, our approach enjoys use of knowledge obtained from denoising as well as painless application to downstream tasks with no access to accurate conformers. Experiments on real-world molecular property prediction datasets show that the graph encoder trained via \ourmethod{} can infer 3D information based on the 2D graph and shows superior performance and label-efficiency against other baselines.
\end{abstract}

\section{Introduction}\label{sec:introduction}
\cutsubsectiondown


Molecular property prediction has gained much interest across the machine learning community, leading to breakthroughs in various applications such as drug discovery~\cite{GUVENCH20161928, https://doi.org/10.1111/cbdd.12952} and material design~\cite{doi:10.1146/annurev-matsci-082019-105100, https://doi.org/10.1002/adfm.201501919, 2019npjCM...5...83S, 2022npjCM...8...84P}. As molecules can be represented as a \textit{2D graph} with nodes and edges representing atoms and covalent bonds, many graph neural networks have been developed with promising results~\cite{https://doi.org/10.48550/arxiv.1509.09292, https://doi.org/10.48550/arxiv.1606.09375, https://doi.org/10.48550/arxiv.1312.6203, doi:10.1021/acs.jcim.6b00601, 4700287, https://doi.org/10.48550/arxiv.1812.01070}. However, achieving high precision requires accurate ground-truth property labels which are very expensive to obtain. 
This limitation has motivated adaptation of self-supervised pretraining widely used in natural language processing~\cite{devlin2018bert, brown2020language} and computer vision~\cite{he2020momentum,bachmann2022multimae} onto molecular graphs with proxy objectives developed to instill useful knowledge into neural networks with unlabeled data.
But existing 2D graph-based pretraining frameworks face a fundamental challenge: while the model is trained to learn representations that are invariant 
under various data augmentations, augmenting 2D graphs can catastrophically disrupt its topology, which renders the model unable to fully recover labels from augmented samples~\cite{trivedi2022analyzing}. As a result of this limitation, recent work has shown that existing 2D pretraining approaches do not show statistically meaningful performance improvements in downstream tasks~\cite{sun2022does}. 

\begin{figure}[t]
    \centering 
    \includegraphics[width=\textwidth,trim={0 3.65cm 0 0},clip]{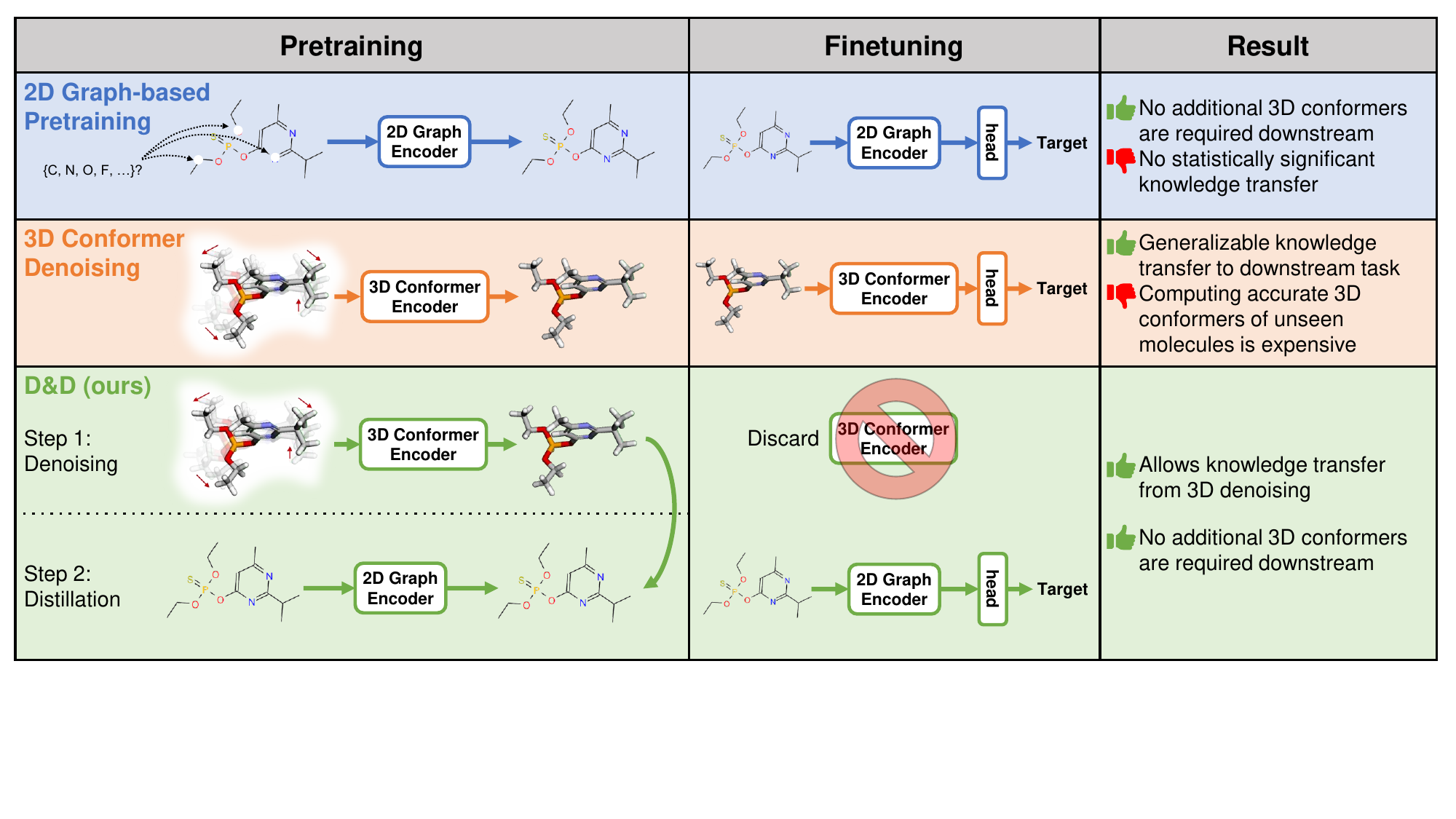}
    \vspace{-.15in}
    \caption{Comparison between \ourmethod{} and existing molecular pretraining frameworks. \textbf{Top:} 2D graph-based pretraining methods fail to bring significant benefit to downstream molecular property prediction. \textbf{Middle:} 3D denoising is effective in predicting molecular properties by approximately learning the force field in the physical space, but cannot be easily applied to downstream tasks where only 2D graphs are available. \textbf{Bottom:} Our method \ourmethod{} allows practitioners to leverage knowledge from 3D denoising in downstream scenarios where only 2D molecular graphs are available without the need to generate 3D conformer via expensive computations or machine learning approaches.}
    \label{figure:introduction_figure}
    \vspace{-.2in}
\end{figure}

As an alternative, recent work have proposed incorporating 3D information to the pretraining objective, leveraging large unlabeled datasets of \textit{3D conformers}, or point clouds of atoms floating in the physical space.
While a natural task would be to reconstruct the input conformer, this may not induce generalizable knowledge as each conformer only represents a single local minima in a distribution of 3D configurations.
On the other hand, the force field that controls the overall stabilization process provides significant chemical information that can be used across many different molecular properties~\cite{mezey2001distributions}.
This naturally translates to pretraining via denoising conformers under 
perturbations, an approach that has shown state-of-the-art performance in diverse molecular property prediction benchmarks~\cite{zaidi2022pvd, liu2022molecular}. 

Despite great performance, a model trained with denoising requires conformers downstream as well, and obtaining accurate conformers require expensive quantum mechanical computations. While there exist many rule-based~\cite{riniker2015etkdg, landrum2016rdkit} as well as deep learning-based approaches~\cite{ganea2021geomol, xu2022geodiff, jing2022torsional} for generating conformers, previous work have shown that existing methods fail to generate conformers quickly and accurately enough to be used in a large scale~\cite{stark20223dinfomax}.

    
In light of such limitations, we propose \ourmethod{} (Denoise and Distill), a self-supervised molecular representation learning framework that enjoys the best of both worlds. Figure~\ref{figure:introduction_figure} shows the overall pipeline of our work. \ourmethod{} sequentially performs two steps: 1) we pretrain a 3D teacher model that denoises conformers artificially perturbed with Gaussian noise and 2) freeze the 3D teacher encoder and distill representations from the 3D teacher onto the 2D student. When given a downstream task with access to 2D molecular graphs only, the 3D teacher is discarded and the 2D student is finetuned towards the given task. As a result of distillation, \ourmethod{} encourages the 2D graph encoder to exploit the topology of the molecular graph towards encoding the input molecule similarly to the 3D conformer encoder without any explicit supervision from property labels.
Surprisingly, experiments on various molecular property prediction datasets indicate that the 2D graph representations from \ourmethod{} can generalize to unseen molecules.
To the best of our knowledge, our method is the first self-supervised molecular representation learning framework that adopts cross-modal knowledge distillation to transfer knowledge from a 3D denoiser onto a 2D graph encoder. 
We summarize our main contributions as follows:
\begin{itemize}
    \item We propose \ourmethod{}, a two-step self-supervised molecular representation pretraining framework that performs 3D-to-2D cross-modal distillation.
    \item Pretraining results show that under \ourmethod{}, the 2D student model can closely mimic representations from the 3D teacher model using graph features and topology. 
    Further analysis shows that the intermediate representations of the 2D student also aligns well with 3D geometry.
    \item Experiments on the OGB benchmark and manually curated physical property datasets show that \ourmethod{} leads to significant knowledge transfer, and also performs well in downstream scenarios where the number of labeled training data points is limited.
\end{itemize}

\cutsubsectionup
\section{Related Work}\label{sec:related_work}
\cutsubsectiondown
In this section, we first discuss previous work on knowledge distillation that inspired our approach. We also cover existing self-supervised pretraining approaches for molecular representation learning.

\paragraph{Knowledge Distillation.}

Knowledge distillation (KD) was developed under the motivation of transferring knowledge learned by a large \textit{teacher} model to a much more compact \textit{student} model, thereby reducing the computational burden while preserving the predictive performance~\cite{hinton2015distilling}. Example approaches in computer vision include distilling class probabilities as a soft target for classification models~\cite{ba2014deep} or transferring intermediate representations of input images~\cite{tian2019contrastive}. 
For dense prediction tasks such as semantic segmentation, it has been shown that a \textit{structured} KD approach that distills pixel-level features instead leads to improvements in performance~\cite{liu2020structured}. Another extension that is more closely related to our approach is \textit{cross-modal} KD on unlabeled modality-paired data (e.g. RGB and Depth images), which was proposed to cope with modalities with limited data~\cite{gupta2016cross}. Inspired by this work, \ourmethod{} performs 3D-to-2D cross-modal KD to allow downstream finetuning on 2D molecular graphs while utilizing the feature space refined by 3D conformer denoising. Further information on KD can be found in a recent survey by~\cite{gou2021knowledge}.

\paragraph{Pretraining for Molecular Property Prediction.}

Inspired by previous work in the NLP domain, there exist many self-supervised pretraining approaches for learning representations of molecular graphs. 
Similar to masked token prediction in BERT~\cite{devlin2018bert}, \cite{hu2019strategies} proposed
node-attribute masking and context prediction to reconstruct topological structures or predict attributes of masked nodes. 
GROVER~\cite{rong2020self} proposed predicting what motifs exist in the molecular graph, under the insight that
functional groups determine molecular properties. Contrastive approaches were also proposed, in which the task is to align representations from two augmentations of the same molecule and repel representations of different ones~\cite{hassani2020contrastive, you2020graphcl}. Despite promising results, it has been shown that obtaining significant gains in performance with existing 2D pretraining methods is non-trivial, as empirical improvements rely heavily on the choice of hyperparameters and other experimental setups~\cite{sun2022does}.

As molecules lie in the 3D physical space, some work have deviated from the 2D graph setting and instead proposed 3D pretraining via denoising conformers perturbed with Gaussian noise, from which empirical results have shown significant knowledge transfer to diverse molecular property prediction tasks~\cite{zaidi2022pvd, liu2022molecular}. Despite great downstream performance, such 3D approaches necessitate access to accurate 3D conformers of molecules under concern, which are difficult to obtain as it requires expensive quantum mechanical computations such as density functional theory (DFT)~\cite{parr1995dft}.

There exist solutions to avoid these drawbacks. 3DInfomax~\cite{stark20223dinfomax} proposed cross-modal contrastive pretraining frameworks that align 2D and 3D representations. In addition to contrastive pretraining, GraphMVP~\cite{liu2021graphmvp} also incorporates generative pretraining which trains the model to reconstruct the 2D graph representation from its 3D counterpart, and vice versa. 
While these methods use 3D conformers during pretraining only, 
they do not capture information from molecular force fields, which we conjecture to be helpful for forward knowledge transfer.
Our \ourmethod{} framework, on the other hand, enjoys the same advantages while also leveraging generalizable knowledge obtained through conformer denoising. 

\cutsubsectionup
\section{Preliminaries}\label{sec:preliminaries}
\cutsubsectiondown
We first introduce preliminary information on learning representations of 2D molecular graphs and 3D molecular conformers alongside notations that we use in later sections.

\begin{figure}[!t]
    \centering 
    \includegraphics[width=.99\textwidth,trim={4cm 2.8cm 5cm 3cm},clip]{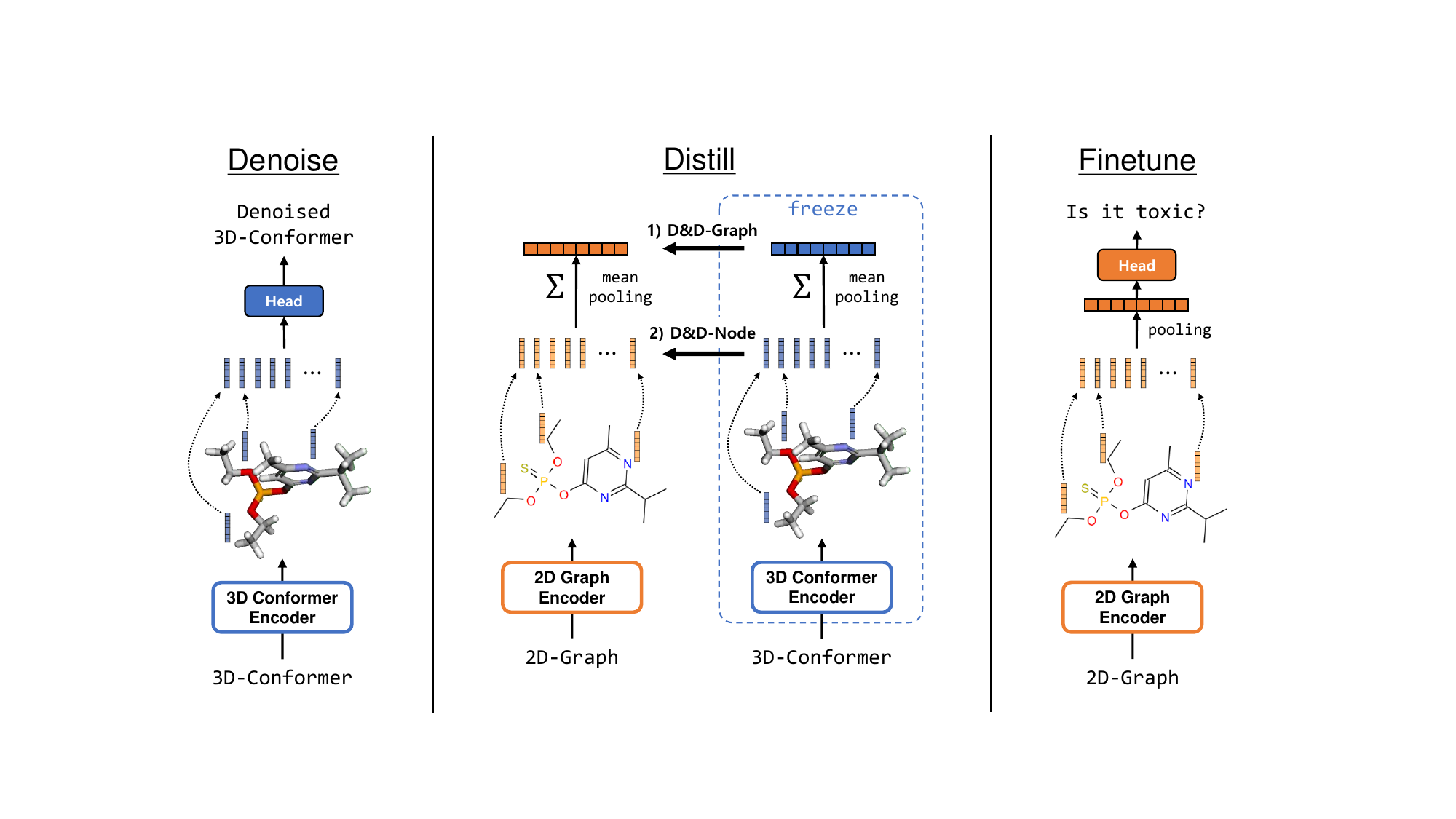}
    \vspace{-.0in}
    \caption{Illustration of our \ourmethod{} framework. First we pretrain a 3D conformer encoding module by denoising perturbed conformers. Next we pretrain a 2D graph encoder by distilling representations from the 3D teacher. We propose two variants: \textsc{\ourmethod{}-Graph} distills mean-pooled graph representations while \textsc{\ourmethod{}-Node} distills node representations in a more fine-grained manner. During finetuning, we tune the 2D graph encoder only with the given downstream data.}
    \label{figure:main_method}
    \vspace{-.1in}
\end{figure}

\paragraph{2D Molecular Graphs.} Let $\mathcal{G}=(\mathcal{V}, \mathcal{E})$ denote the 2D molecular graph with $N$ atoms represented as nodes in $\mathcal{V}$, and $M$ bonds represented as edges in $\mathcal{E}$. In addition to the graph-connectivity, each node is assigned features based on chemical attributes
such as atomic number and aromaticity, and similarly for each edge with features based on bond type and stereo configurations. Given the graph $\mathcal{G}$, a 2D graph encoder $f^{\text{2D}}$ typically first returns representations for each node:
\begin{align}
f^{\text{2D}}(\mathcal{G}) = \BZ^{\text{2D}} \text{ where } \BZ^{\text{2D}} \in \R^{N\times d}.
\end{align} 
In molecular property prediction settings we need a single representation for each molecular graph. 
Typical operators used to extract graph-level representations include mean-pooling all node representations or adding a virtual node to the input graph and treating its representation as the graph representation~\cite{hamilton2020graph}. 

\paragraph{3D Molecular Conformers.} Each molecule can also be represented as a 3D conformer $\mathcal{C}=(\mathcal{V}, \BR)$ with 3D spatial coordinates of each atom stored in $\BR\in \R^{N\times 3}$. Note that unlike in 2D graphs, 3D conformers have information on the graph connectivity nor covalent bonds in $\mathcal{E}$, and is instead treated as point cloud data. As with 2D graphs, let $f^{\text{3D}}$ denote the 3D conformer encoder that takes the conformer $\mathcal{C}$ and returns representations of each atom:
\begin{align}
f^{\text{3D}}(\mathcal{C}) = \BZ^{\text{3D}} \text{ where } \BZ^{\text{3D}} \in \R^{N \times d}.
\end{align}
Note that how the molecule is oriented in the 3D space naturally does not affect its chemical property. Thus, the encoder $f_{3D}$ must return representations that are invariant under
rotations and translations on $\BR$ (i.e. $f^{\text{3D}}((\mathcal{V}, \BR)) = f^{\text{3D}}((\mathcal{V}, g(\BR))$ for $g \in SE(3)$) for efficient weight-tying across SE(3) roto-translations. Since molecular properties are not invariant to chiral orientations, we only respect rotations and translations, but not reflections. There exist many architectures that respect SE(3) symmetry as an inductive bias~\cite{fuchs2020se, bronstein2021geometric, gasteiger2021gemnet, satorras2021egnn, tholke2022torchmd}, and any such architecture can be used for $f^{\text{3D}}$.

\cutsubsectionup
\section{\ourmethod{}: Denoise and Distill}\label{sec:method}
\cutsubsectiondown
Here we describe \ourmethod{}, a molecular pretraining framework that transfers generalizable knowledge from 3D conformer denoising to a 2D graph encoder via cross-modal distillation, thereby allowing painless downstream applications without computing accurate conformers of unseen graphs. The two major steps are as follows: 1) Denoising perturbed conformers with a 3D conformer encoder $f^{3D}$, and 2) Distilling representations from the 3D teacher to the 2D graph encoder $f^{2D}$. An illustration of the overall pipeline can be found in Figure~\ref{figure:main_method}. As our first step of \ourmethod{} is based upon previous work on conformer denoising~\cite{zaidi2022pvd,liu2022structured}, we provide a brief outline of the task and refer readers to corresponding papers for further details and theoretical implications.

\paragraph{Step 1: Pretraining via denoising.} Given a stabilized ground-truth conformer $\mathcal{C} = (\mathcal{V}, \BR)$, $f^\text{3D}$ is given as input a perturbed version of the same conformer $\tilde{C} = (\mathcal{V}, \tilde{\BR})$, produced by slightly perturbing the coordinates of each atom with Gaussian noise as
\begin{align}
    \tilde{\BR}_i = \BR_i + \sigma \Bepsilon_i \text{ where } \Bepsilon_i \sim \mathcal{N}(0, \BI_3)
\end{align}
with noise scale $\sigma$ as hyperparameter. Then, we attach a prediction head $h^{\text{3D}}: \R^{N \times d} \to \R^{N \times 3}$ to the $f^{\text{3D}}$ such that the combined model outputs 3-dimensional vectors per atom.
\begin{align}
    h^{\text{3D}}(f^{\text{3D}}(\tilde{\mathcal{C}})) = (\hat{\Bepsilon}_1, \dots, \hat{\Bepsilon}_N)
\end{align}
Lastly, the model is trained to predict the noise that has been injected to create $\tilde{\mathcal{C}}$ from $\mathcal{C}$. The denoising loss minimized during training is as follows: 
\begin{align}
    \loss_{\text{denoise}} = \E_{p(\tilde{\mathcal{C}},\mathcal{C})} \left[\left\|h^{\text{3D}}(f^{\text{3D}}(\mathcal{\tilde{C}}))-(\Bepsilon_1,\dots, \Bepsilon_N)\right\|_2^2\right]
\end{align}
where $p(\tilde{\mathcal{C}}, \mathcal{C})$ denotes the probability distribution induced by the dataset distribution and the noise sampling procedure to create $\tilde{\mathcal{C}}$. Surprisingly, the denoising objective is equivalent to learning an approximation of the actual force field in the physical space derived by replacing the true distribution of conformers with a mixture of Gaussians~\cite{zaidi2022pvd}. The Gaussian mixture potential corresponds to the classical harmonic oscillator potential in physics, which is a great approximation scheme for linearized equations such as denoising.

For our experiments, we use the TorchMD-NET~\cite{tholke2022torchmd} architecture for $f^{\text{3D}}$ following \citealt{zaidi2022pvd} due to its equivariance to SE(3) roto-translations and high performance on quantum mechanical property prediction. Note that \ourmethod{} is architecture-agnostic, and any other SE(3)-equivariant architecture can be used as well.

\paragraph{Step 2: Cross-modal distillation.} After pretraining via denoising is done, we distill representations from the pretrained $f^{\text{3D}}$ to a 2D graph encoder model $f^{\text{2D}}$. We consider two different variants of cross-modal KD, leading to two respective variants of our approach. For the first variant \textsc{\ourmethod{}-Graph}, we minimize the difference between graph representations from 2D and 3D encoders:

\begin{align}
    \loss_{\text{distill-graph}} = \left\| \text{pool}(f^{\text{2D}}(\mathcal{G})) - \text{pool}(f^{\text{3D}}(\mathcal{C}))\right\|_2^2
\end{align}

During training, we freeze the teacher model $f^{\text{3D}}$ and flow gradients only through the student model $f^{\text{2D}}$. This effectively trains the 2D encoder to leverage the bond features and graph topology to imitate representations from 3D conformers. To obtain graph representations, we average all node representations inferred by each encoder.

Inspired by structured KD~\cite{liu2020structured}, we propose another variant \textsc{\ourmethod{}-Node} that distillation node-level representations without any pooling:

\begin{align}
    \loss_{\text{distill-node}} = \left\| f^{\text{2D}}(\mathcal{G}) - f^{\text{3D}}(\mathcal{C})\right\|_2^2
\end{align}

Unlike \textsc{\ourmethod{}-Graph}, \textsc{\ourmethod{}-Node} makes full use of the one-to-one correspondence between atoms in the molecular graph
and atoms in the conformer. Hence $f^{\text{2D}}$ is trained to align towards representations from $f^{\text{3D}}$ in a more fine-grained manner. 

For the $f^{\text{2D}}$, we use the TokenGT architecture
that theoretically enjoys maximal expressiveness across all possible permutation-equivariant operators on 2D graphs ~\cite{kim2022tokengt}. Due to this flexibility, we expect $f^{\text{2D}}$ to be trained to align representations from $f^{\text{3D}}$ as closely as possible, by which we hope to see the effect of distillation to the fullest extent. Furthermore, using an attention-based architecture also allows analysis on the relationship between the attention scores of atom-pairs and their physical distance in the 3D space. Results from which are discussed later in Section~\ref{sec:experiments}. Note that similarly with $f^{\text{3D}}$, however, any other permutation-equivariant graph neural network architecture can be adopted seamlessly.

\paragraph{Downstream finetuning.}

Assuming the downstream task does not provide accurate conformers as input, we discard $f^{\text{3D}}$ after the distillation step and finetune $f^{\text{2D}}$ with molecular graphs only. We use L1 loss and BCE loss for regression and binary-classification tasks, respectively, following previous work~\cite{stark20223dinfomax}. 

Note that we finetune the entire $f^{\text{2D}}$ model instead of just the newly attached prediction head on the downstream data.  Given that the force fields induced by electron clouds provide knowledge that is generalizable to various molecular properties, we conjecture that 
\ourmethod{} provides a good initial point in the parameter space from which finetuning $f^{\text{2D}}$ entirely leads to a better local optima. This also aligns with previous observations in NLP that pretrained language models outperform models trained from scratch only when the entire model is finetuned~\cite{rothermel2021don}.

\section{Experiments}\label{sec:experiments}
\cutsubsectiondown

For empirical evaluation, we test our \ourmethod{} pipeline on various molecular property prediction tasks using open-source benchmarks as well as four manually curated datasets. We also stress-test \ourmethod{} under a downstream scenario where the number of labeled data points is extremely limited. All experiments are run in a remote GCP server equipped with 16 NVIDIA A100 Tensor Core GPUs.

\begin{wrapfigure}{R}{0.5\textwidth}
    \centering 
    \vspace{-.2in}
    \begin{subfigure}[b]{0.245\textwidth}
        \centering
        \includegraphics[width=\textwidth]{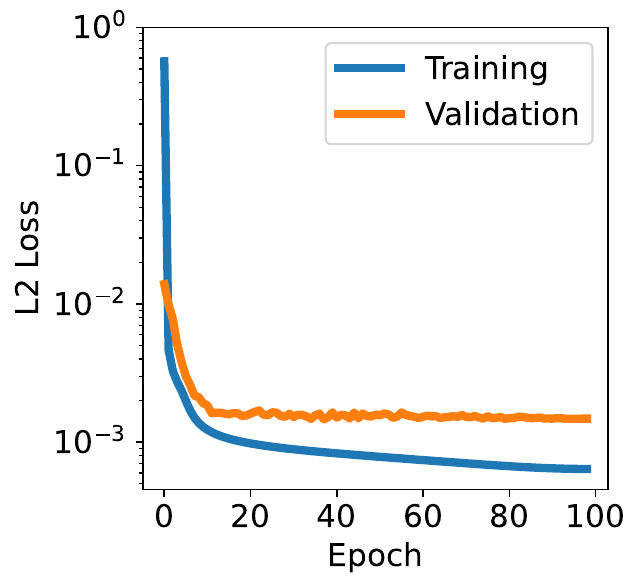}
        \vspace{-.25in}
        \caption{\textsc{\ourmethod{}-Graph}}
    \end{subfigure}\hfill
    \begin{subfigure}[b]{0.245\textwidth}
        \centering
        \includegraphics[width=\textwidth]{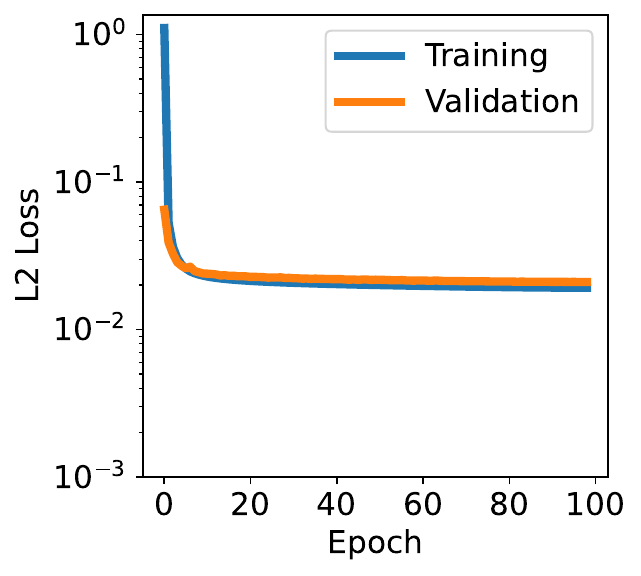}
        \vspace{-.25in}
        \caption{\textsc{\ourmethod{}-Node}}
    \end{subfigure}
    \vspace{-.05in}
    \caption{Training and validation loss curves during distillation of \textsc{\ourmethod{}-Graph} and \textsc{\ourmethod{}-Node} on PCQM4Mv2. All plots are in log-scale. The 2D student is able to closely distill representations from the 3D teacher with small generalization gap.}\label{figure:distillation_curves}
    \vspace{-.1in}
\end{wrapfigure}

\subsection{Experimental Setup} 

\paragraph{Datasets.} For pretraining, we use PCQM4Mv2~\cite{nakata2017pcqm4mv2}, a large molecule dataset consisted of 3.7M molecules. Each molecule is paired with a single 3D conformer at the lowest-energy state computed via DFT. In case of \ourmethod{}, we use the same PCQM4Mv2 dataset for both denoising and disillation steps. Note that even though PCQM4Mv2 provides the HOMO-LUMO energy gap of each molecule as labels, we do not use any supervision from such labels during training, and instead treat the dataset as a collection of unlabeled molecular graph-conformer pairs.

For finetuning, we use ten benchmark datasets published in OGB~\cite{hu2020ogb}, three of which are regression tasks and the rest are binary classification tasks. We use the same scaffold split provided by the OGB library. As shown in Appendix~\ref{app:dataset}, the OGB datasets exhibit different molecule distributions from PCQM4Mv2: some tasks involve atom types that the encoder has never observed during pretraining. 
We also manually curate four datasets on physical molecular properties:
Melting point (MP) is a phase transition temperature from solid to liquid \cite{Bradley2014}. Boiling point (BP) is also a phase transition temperature from liquid to gas \cite{10.1093/nar/gkac956}. Refractive index (RI) measures the relative speed of light in medium to a vacuum \cite{YAWS20151}. LogP is the partition coefficient which indicates the ratio of concentrations of a compound in a mixture of two immiscible solvents, water and octanol, at equilibrium \cite{10.1093/nar/gkac956}. These four datasets are normalized by mean and standard deviation before training and evaluation. We randomly split the dataset 8:1:1 for training, validation and testing, respectively. The detailed dataset statistics can be found in Appendix~\ref{app:dataset}.

\paragraph{Baselines.}

We compare \textsc{\ourmethod{}-Graph} and \textsc{\ourmethod{}-Node} against two baselines. \textsc{RandInit} is a na\"ive baseline that trains $f^{\text{2D}}$ on each downstream task starting from randomly initialized model weights. 3DInfomax~\cite{stark20223dinfomax} is a contrastive pretraining approach that considers two representations, one from $f^{\text{2D}}$ and another from $f^{\text{3D}}$, to be a positive pair if they result from the same molecule, or negative pair otherwise. Given a batch of molecules, it minimizes the NTXent loss introduced in SimCLR~\cite{chen2020simclr} to align the positive pairs together and repel the negative pairs within the feature space.
For 3DInfomax, we use cosine similarity for the similarity function $sim(\cdot,\cdot)$ and temperature $\tau = 0.01$ for weighting negative pairs as suggested in the original paper~\cite{stark20223dinfomax}.
    


When finetuning, we consider two pooling operators for extracting graph representations: 1) mean-pooling all node representations (\textsc{+mp}) and 2) using the virtual-node representation as the graph representation (\textsc{+vn}). For consistency, we follow the same featurization step provided by the OGB library across all experiments, which produces a 9 and 3-dimensional feature vector for each atom and bond, respectively. For reproducibility, we provide the list of hyperparameters in Appendix~\ref{app:hyperparam}.

\cutsubsectiondown
\subsection{Pretraining Results}
\cutsubsectiondown

Prior to downstream evaluation, we discuss interesting findings from pretraining with \ourmethod{}.

\paragraph{The 2D graph encoder can closely mimic representations from 3D conformers using only the molecular graph.} Figure~\ref{figure:distillation_curves} shows the training and validation loss curves of the distillation step of \ourmethod{}. When pretraining with \textsc{\ourmethod{}-Node}, the distillation loss converges to slightly over $10^{-2}$ with a very small generalization gap between validation and training. This shows that the 2D molecular graph contains enough information to closely imitate representations from the 3D teacher $f^{\text{3D}}$. The small gap between training and validation also reflects that the guidance provided via \textsc{\ourmethod{}-Node} can well-generalize towards unseen molecules. When pretraining with \textsc{\ourmethod{}-Graph}, we find that the training loss converges to a much lower optima of $10^{-3}$, but with a much larger generalization gap of approximately $10^{-3}$. This implies that while the task of distilling mean-pooled representations is easier than distilling node-wise representations, it leads to less generalizable knowledge due to not considering the graph topological structure.

\begin{figure}[!t]
    \centering 
    \begin{subfigure}[b]{0.31\textwidth}
        \centering
        \includegraphics[width=\textwidth]{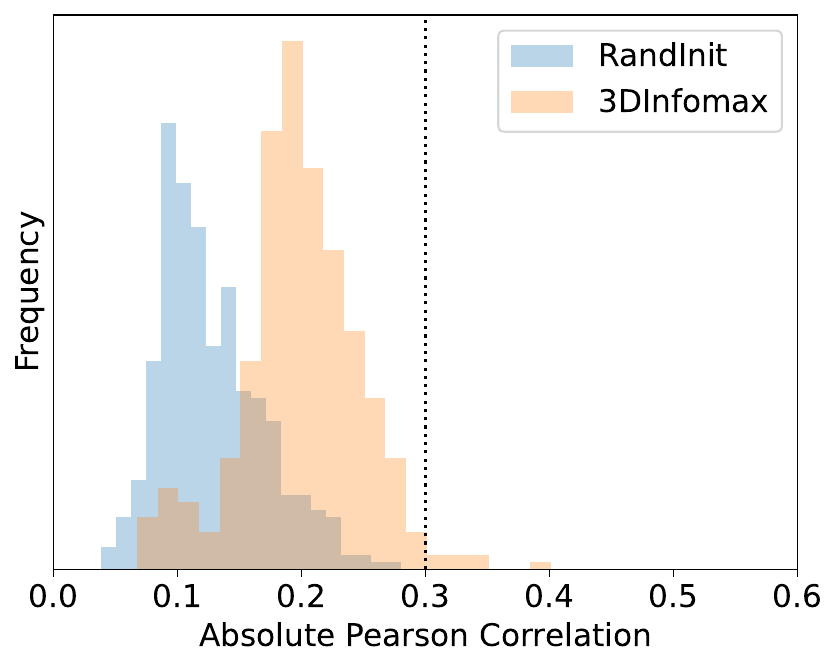}
        \vspace{-.15in}
    \end{subfigure}
    \hfill
    \begin{subfigure}[b]{0.31\textwidth}
        \centering
        \includegraphics[width=\textwidth]{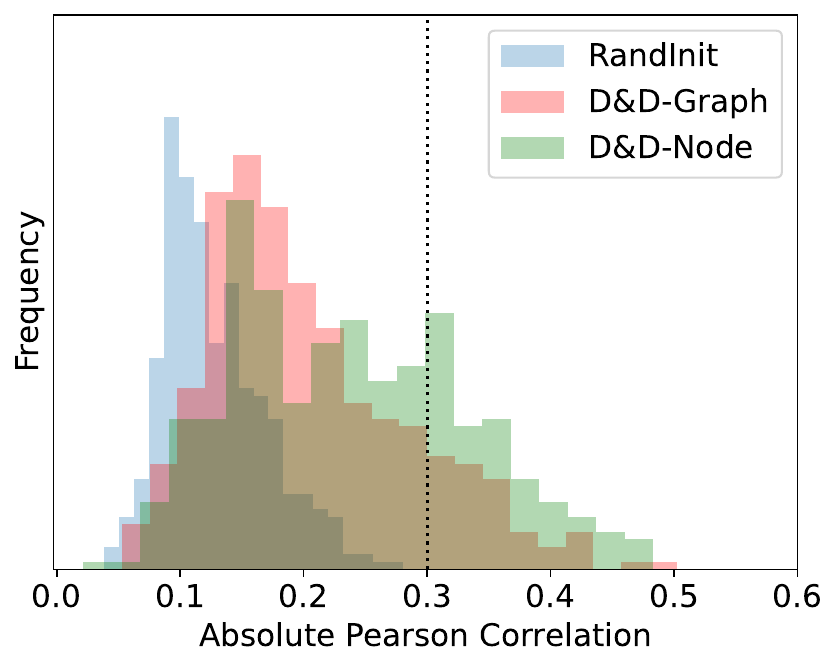}
        \vspace{-.15in}
    \end{subfigure}
    \hfill
    \begin{subfigure}[b]{0.32\textwidth}
        \centering
        \includegraphics[width=\textwidth]{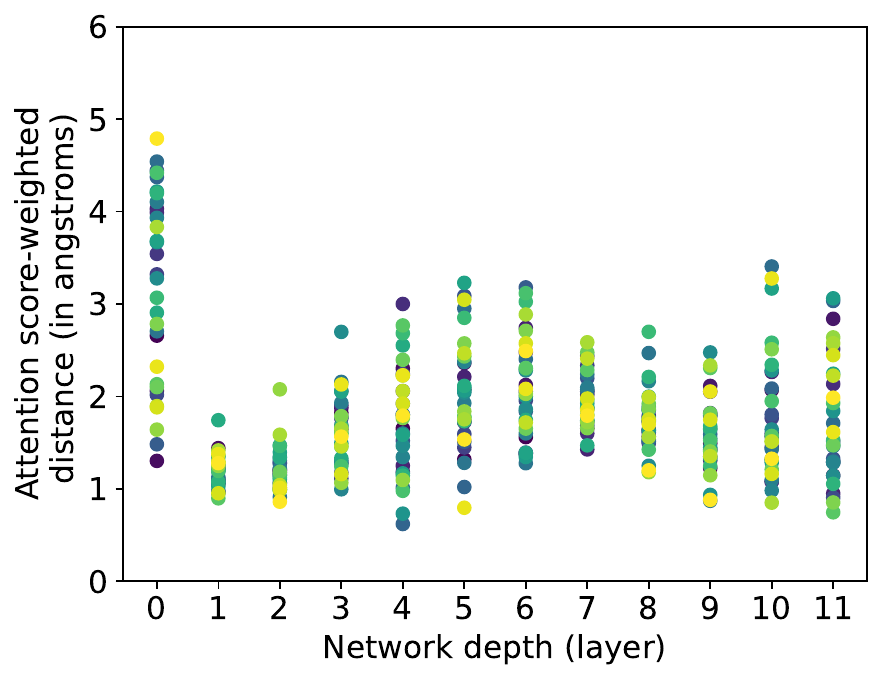}
        \vspace{-.15in}
    \end{subfigure}
    \vspace{-.1in}
    \caption{Histograms of Pearson correlation values between pre-softmax attention scores vs. 3D pairwise distance during inference on the PCQM4Mv2 validation set for (Left) \textsc{Contrastive}, (Middle) \textsc{\ourmethod{}-Graph} and \textsc{\ourmethod{}-Node}. (Right) Average attention score-weighted 3D distances according to network depth from \textsc{\ourmethod{}-Node}. Each colored dot represents an attention head in the corresponding layer.}
    \label{fig:attention_scores}
    \vspace{-.2in}
\end{figure}

\paragraph{The intermediate encoding procedure of the 2D encoder trained via \ourmethod{} aligns with 3D geometry.}
As we use an attention-based architecture for $f^{\text{2D}}$, we qualitatively assess whether the encoder processes molecular graphs as if 3D geometry without ground-truth conformers by evaluating how it performs attention across atoms during inference (e.g. do atoms nearby in the 3D space tend to attend to each other?). Specifically, we compute the absolute Pearson correlation between the 3D pairwise distances of atoms and the inner product of their features prior to the softmax layer in each attention head, averaged across all molecules in the PCQM4Mv2 validation set. Note that a larger inner product implies a relatively larger exchange of information between the two atoms. The first two figures in Figure~\ref{fig:attention_scores} show histograms depicting distributions of averaged absolute Pearson correlation values from all attention heads for each layer in the 2D encoder after pretraining by each method. We find that 3DInfomax only leads to a slight increase in correlation compared to \textsc{RandInit}: most correlation values are distributed under 0.3. When pretrained with our \ourmethod{}-variants, however, many attention heads show correlations that exceed 0.3, a value that is never reached with randomly initialized weights. This implies that our approach provides guidance to the 2D graph encoder towards processing molecular graphs while respecting its 3D geometry.

For further investigation, we also measure the average pairwise distances weighted by the attention scores from \textsc{\ourmethod{}-Node} with results shown in the rightmost plot in Figure~\ref{fig:attention_scores}. A higher value indicates that the attention head tends to exchange information across atoms that are far apart. Interestingly, the first layer exhibits a diverse range of distances, but the 
layer that immediately follows uses attention mostly to exchange information across atoms that are geometrically nearby each other, similar to a SE(3)-convolutional layer. Considering that a carbon-carbon single bond has an average length of 1.5 angstroms, this result indicates that $f^{\text{2D}}$ pretrained with \textsc{\ourmethod{}-Node} can reason about 3D geometry to exchange information across atoms that are nearby in the 3D conformer, even though they may be far apart in the 2D graph.


\cutsubsectiondown
\subsection{Finetuning Results}
\cutsubsectiondown

Here we first provide empirical observations from downstream evaluation on OGB and Curated datasets. We also experiment on the QM9 dataset using a different 2D GNN model, to show that our pretraining approach is effective with other architectures as well.

\begin{table}[!t]
    \centering
    \caption{Average performance and standard deviations for OGB and Curated datasets: 12 tasks on the left are from OGB, and 4 tasks on the right are from manually curated data. Best results are in \textbf{bold}.}\label{tab:main_results} 
    \vspace{.05in}
    \resizebox{\textwidth}{!}{\begin{tabular}{rr|ccccc|cc}
        \toprule
        & & \multicolumn{5}{c|}{OGB} & \multicolumn{2}{c}{Curated}\\
        \midrule
        \multicolumn{2}{r|}{Dataset} & ESOL & FREESOLV & LIPO & BACE & BBBP & BP & MP\\
        \multicolumn{2}{r|}{Metric} & RMSE($\downarrow$) & RMSE($\downarrow$) & RMSE($\downarrow$) & ROC-AUC($\uparrow$) & ROC-AUC($\uparrow$) & MAE($\downarrow$) & MAE($\downarrow$)\\
        \midrule
        \multirow{2}*{\textsc{RandInit}} & \textsc{+mp} & 1.0491{\scriptsize $\pm$0.0363} & 3.6445{\scriptsize $\pm$0.0734} & 0.9720{\scriptsize $\pm$0.0185} & 0.6360{\scriptsize $\pm$0.1055} & 0.6453{\scriptsize $\pm$0.0153} & 0.3000{\scriptsize $\pm$0.0027} & 0.3672{\scriptsize $\pm$0.0049} \\
        & \textsc{+vn} & 1.1074{\scriptsize $\pm$0.0735} & 3.5325{\scriptsize $\pm$0.0453} & 0.9703{\scriptsize $\pm$0.0145} & 0.7078{\scriptsize $\pm$0.0284} & 0.6305{\scriptsize $\pm$0.0066} & 0.3014{\scriptsize $\pm$0.0052} & 0.3662{\scriptsize $\pm$0.0033} \\
        \multirow{2}*{\textsc{3DInfomax}} & \textsc{+mp} & 2.0037{\scriptsize $\pm$0.1076} & 4.6854{\scriptsize $\pm$0.4345} & 1.0167{\scriptsize $\pm$0.0214} & 0.5701{\scriptsize $\pm$0.0924} & 0.6238{\scriptsize $\pm$0.0120} & 0.3257{\scriptsize $\pm$0.0077} & 0.3862{\scriptsize $\pm$0.0016}\\
        & \textsc{+vn} & 1.5246{\scriptsize $\pm$0.1092} & 4.2580{\scriptsize $\pm$0.3782} & 0.9600{\scriptsize $\pm$0.0206} & 0.6016{\scriptsize $\pm$0.4050} & 0.6358{\scriptsize $\pm$0.0088} & 0.3160{\scriptsize $\pm$0.0053} & 0.3791{\scriptsize $\pm$0.0022}\\
        \midrule
        \multirow{2}*{\textsc{\ourmethod{}-Graph}} & \textsc{+mp} & \bf 0.9276{\scriptsize $\pm$0.0562} & \bf 2.6814{\scriptsize $\pm$0.0867} & 0.7646{\scriptsize $\pm$0.0126} & 0.6906{\scriptsize $\pm$0.1339} & 0.6775{\scriptsize $\pm$0.0117} & 0.2476{\scriptsize $\pm$0.0063} & 0.3156{\scriptsize $\pm$0.0017}\\
        & \textsc{+vn} & 0.9934{\scriptsize $\pm$0.0203} & 2.7004{\scriptsize $\pm$0.0572} & 0.7533{\scriptsize $\pm$0.0101} & \bf 0.7271{\scriptsize $\pm$0.0721} & 0.6714{\scriptsize $\pm$0.0144} & 0.2389{\scriptsize $\pm$0.0018} & 0.3096{\scriptsize $\pm$0.0023}\\
        \multirow{2}*{\textsc{\ourmethod{}-Node}} & \textsc{+mp} & 1.0097{\scriptsize $\pm$0.0381} & 3.2935{\scriptsize $\pm$0.2872} & \bf 0.6975{\scriptsize $\pm$0.0197} & 0.5882{\scriptsize $\pm$0.2827} & 0.6754{\scriptsize $\pm$0.0028} & \bf 0.2220{\scriptsize $\pm$0.0022} & \bf 0.2937{\scriptsize $\pm$0.0003}\\
        & \textsc{+vn} & 1.0862{\scriptsize $\pm$0.0887} & 3.4056{\scriptsize $\pm$0.2221} & 0.7425{\scriptsize $\pm$0.0113} & 0.6649{\scriptsize $\pm$0.1460} & \bf 0.6788{\scriptsize $\pm$0.0068} & 0.2296{\scriptsize $\pm$0.0068} & 0.2965{\scriptsize $\pm$0.0008}\\
        \midrule
        \multicolumn{2}{r|}{Dataset} & CLINTOX & HIV & SIDER & TOX21 & TOXCAST & RI & logP\\
        \multicolumn{2}{r|}{Metric} & ROC-AUC($\uparrow$) & ROC-AUC($\uparrow$) & ROC-AUC($\uparrow$) & ROC-AUC($\uparrow$) & ROC-AUC($\uparrow$) & MAE($\downarrow$) & MAE($\downarrow$)\\
        \midrule
       \multirow{2}*{\textsc{RandInit}} & \textsc{+mp} & 0.6141{\scriptsize $\pm$0.0290} & 0.7291{\scriptsize $\pm$0.0213} & 0.6005{\scriptsize $\pm$0.0058} & 0.7219{\scriptsize $\pm$0.0035} & 0.6275{\scriptsize $\pm$0.0046} & 0.1243{\scriptsize $\pm$0.0032} & 0.0501{\scriptsize $\pm$0.0043}\\
       & \textsc{+vn} & 0.6834{\scriptsize $\pm$0.0468} & 0.6758{\scriptsize $\pm$0.0220} & 0.5930{\scriptsize $\pm$0.0036} & 0.7190{\scriptsize $\pm$0.0051} & 0.6330{\scriptsize $\pm$0.0030} & 0.1250{\scriptsize $\pm$0.0034} & 0.0503{\scriptsize $\pm$0.0034}\\
        \multirow{2}*{\textsc{3DInfomax}} & \textsc{+mp} & \bf 0.6919{\scriptsize $\pm$0.0419} & 0.7295{\scriptsize $\pm$0.0064} & 0.5979{\scriptsize $\pm$0.0086} & 0.6925{\scriptsize $\pm$0.0114} & 0.5824{\scriptsize $\pm$0.0069} & 0.1465{\scriptsize $\pm$0.0006} & 0.0458{\scriptsize $\pm$0.0010}\\
        & \textsc{+vn} & 0.6815{\scriptsize $\pm$0.0361} & 0.7165{\scriptsize $\pm$0.0062} & 0.6004{\scriptsize $\pm$0.0216} & 0.6979{\scriptsize $\pm$0.0066} & 0.5879{\scriptsize $\pm$0.0033} & 0.1344{\scriptsize $\pm$0.0042} & 0.0413{\scriptsize $\pm$0.0020}\\
        \midrule
        \multirow{2}*{\textsc{\ourmethod{}-Graph}} & \textsc{+mp} & 0.6761{\scriptsize $\pm$0.0211} & 0.7766{\scriptsize $\pm$0.0238} & 0.5988{\scriptsize $\pm$0.0049} & 0.7541{\scriptsize $\pm$0.0014} & \bf 0.6478{\scriptsize $\pm$0.0010} & 0.0852{\scriptsize $\pm$0.0013} & 0.0274{\scriptsize $\pm$0.0041}\\
        & \textsc{+vn} & 0.6710{\scriptsize $\pm$0.0429} & \bf 0.7771{\scriptsize $\pm$0.0108} & 0.6117{\scriptsize $\pm$0.0229} & 0.7554{\scriptsize $\pm$0.0003} & 0.6363{\scriptsize $\pm$0.0035} & 0.0806{\scriptsize $\pm$0.0048} & 0.0251{\scriptsize $\pm$0.0010}\\
        \multirow{2}*{\textsc{\ourmethod{}-Node}} & \textsc{+mp} & 0.5825{\scriptsize $\pm$0.0257} & 0.7672{\scriptsize $\pm$0.0162} & 0.6017{\scriptsize $\pm$0.0113} & 0.7549{\scriptsize $\pm$0.0055} & 0.6432{\scriptsize $\pm$0.0038} & 0.0688{\scriptsize $\pm$0.0017} & \bf 0.0220{\scriptsize $\pm$0.0006}\\
        & \textsc{+vn} & 0.6594{\scriptsize $\pm$0.0502} & 0.7645{\scriptsize $\pm$0.0113} & \bf 0.6257{\scriptsize $\pm$0.0106} & \bf 0.7556{\scriptsize $\pm$0.0068} & 0.6421{\scriptsize $\pm$0.0011} & \bf 0.0666{\scriptsize $\pm$0.0014} & 0.0226{\scriptsize $\pm$0.0003}\\
        \bottomrule
    \end{tabular}
    }
    \vspace{-.1in}
\end{table}
\begin{figure*}[!t]
\centering
\begin{subfigure}[b]{0.195\textwidth}
    \centering
    \includegraphics[width=\textwidth]{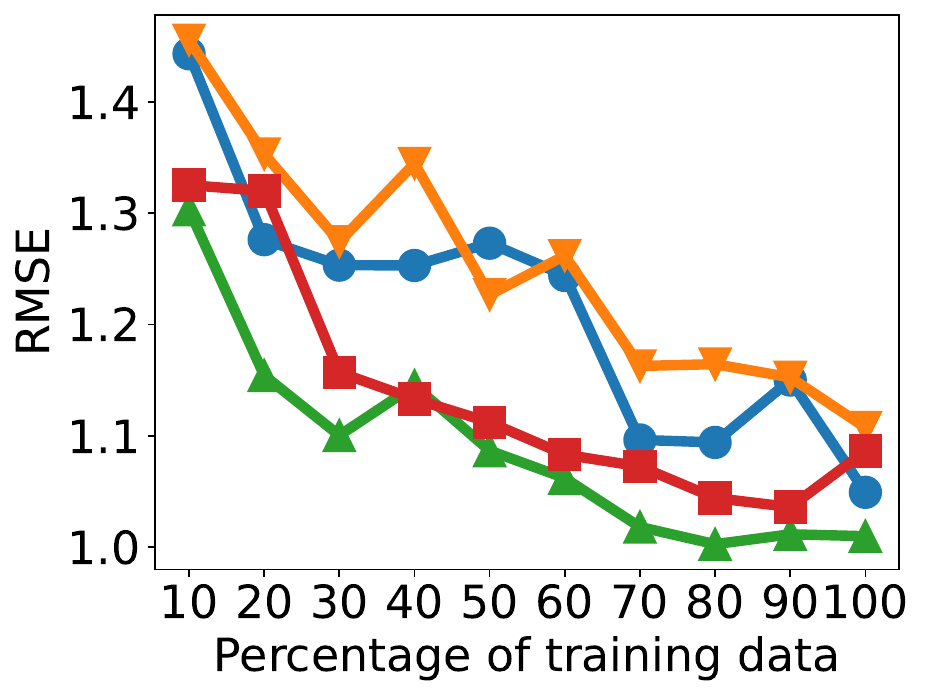}
    \vspace{-.2in}
    \caption{ESOL ($\downarrow$)}  
\end{subfigure}
\hfill
\begin{subfigure}[b]{0.195\textwidth}
    \centering
    \includegraphics[width=\textwidth]{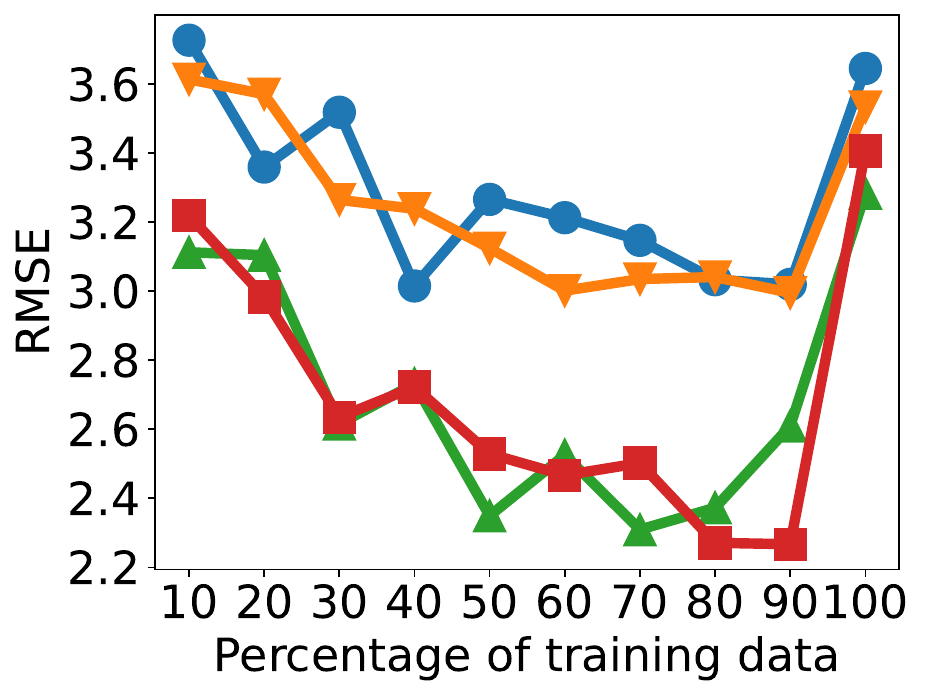}
    \vspace{-.2in}
    \caption{{FREESOLV ($\downarrow$)}}    
\end{subfigure}
\hfill
\begin{subfigure}[b]{0.195\textwidth}
    \centering
    \includegraphics[width=\textwidth]{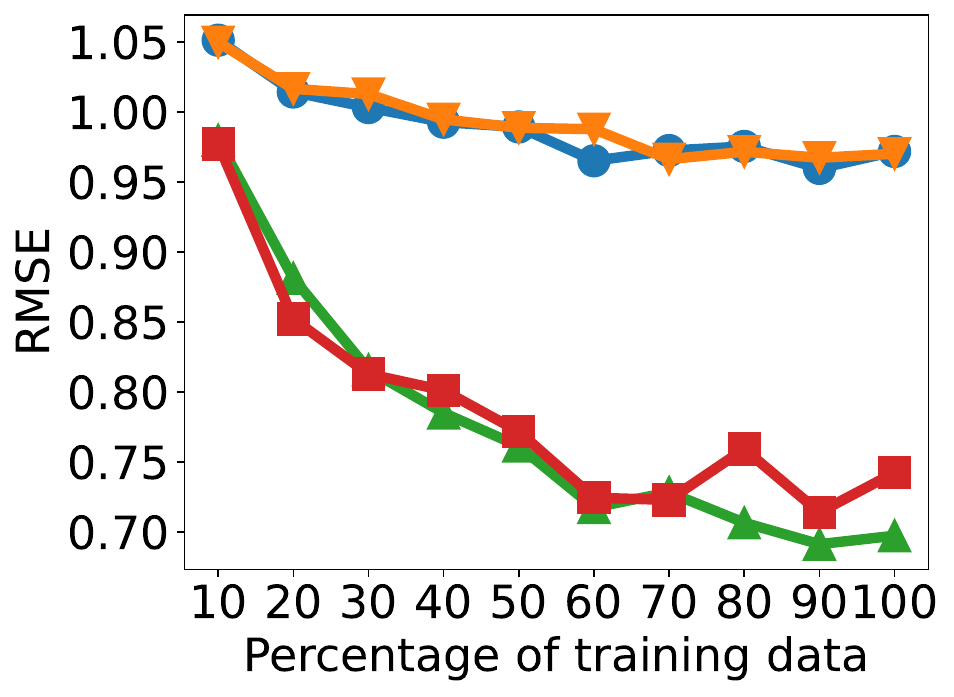}
    \vspace{-.2in}
    \caption{{LIPO ($\downarrow$)}}    
\end{subfigure}
\hfill
\begin{subfigure}[b]{0.195\textwidth}
    \centering
    \includegraphics[width=\textwidth]{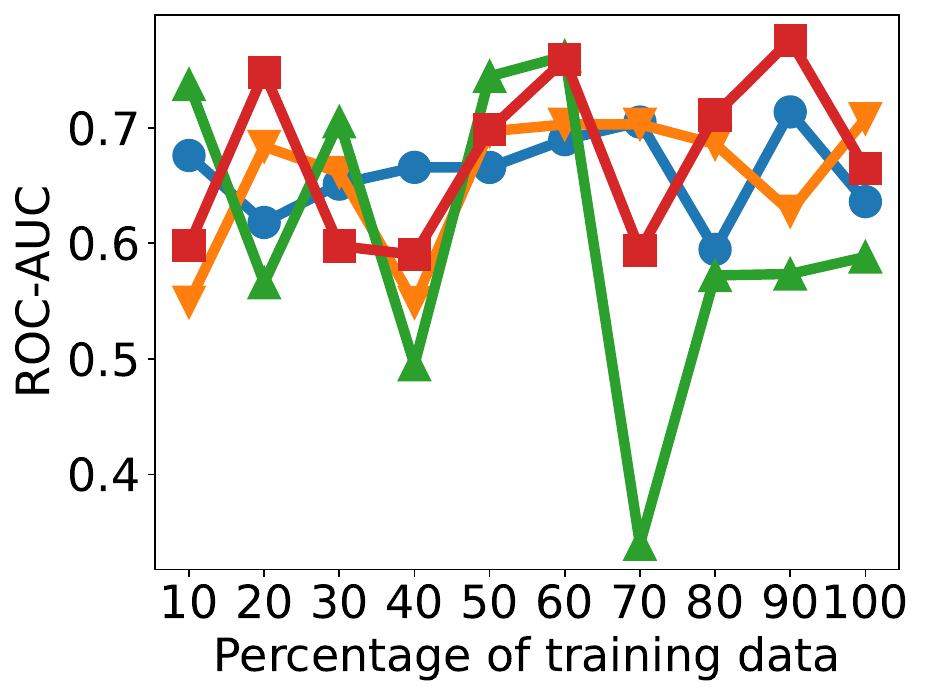}
    \vspace{-.2in}
    \caption{{BACE ($\uparrow$)}}    
\end{subfigure}
\hfill
\begin{subfigure}[b]{0.195\textwidth}
    \centering
    \includegraphics[width=\textwidth]{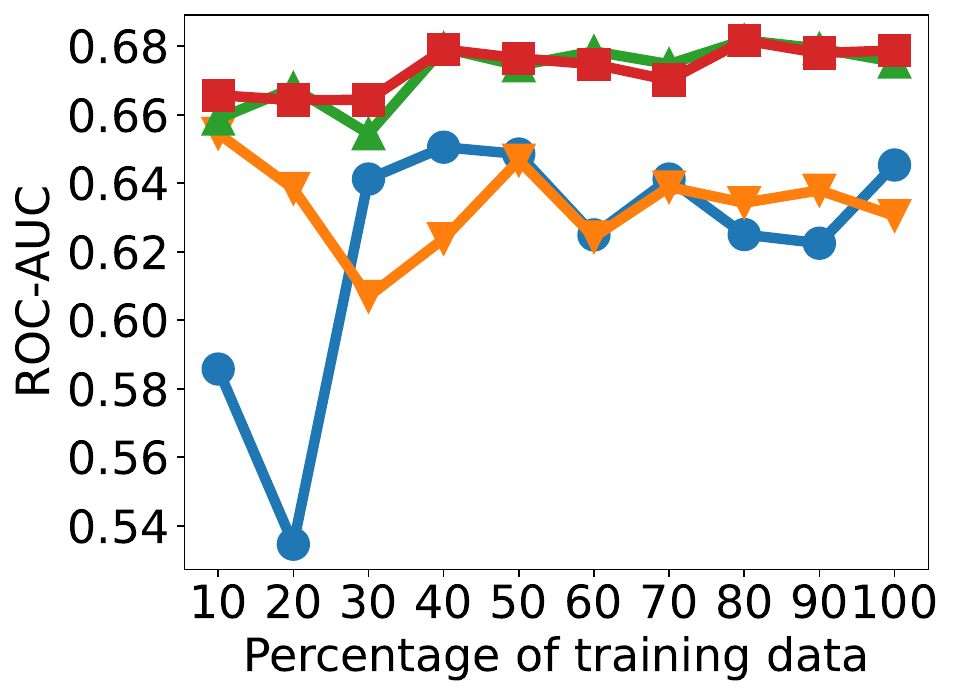}
    \vspace{-.2in}
    \caption{{BBBP ($\uparrow$)}}    
\end{subfigure}
\\
\begin{subfigure}[b]{0.195\textwidth}
    \centering
    \includegraphics[width=\textwidth]{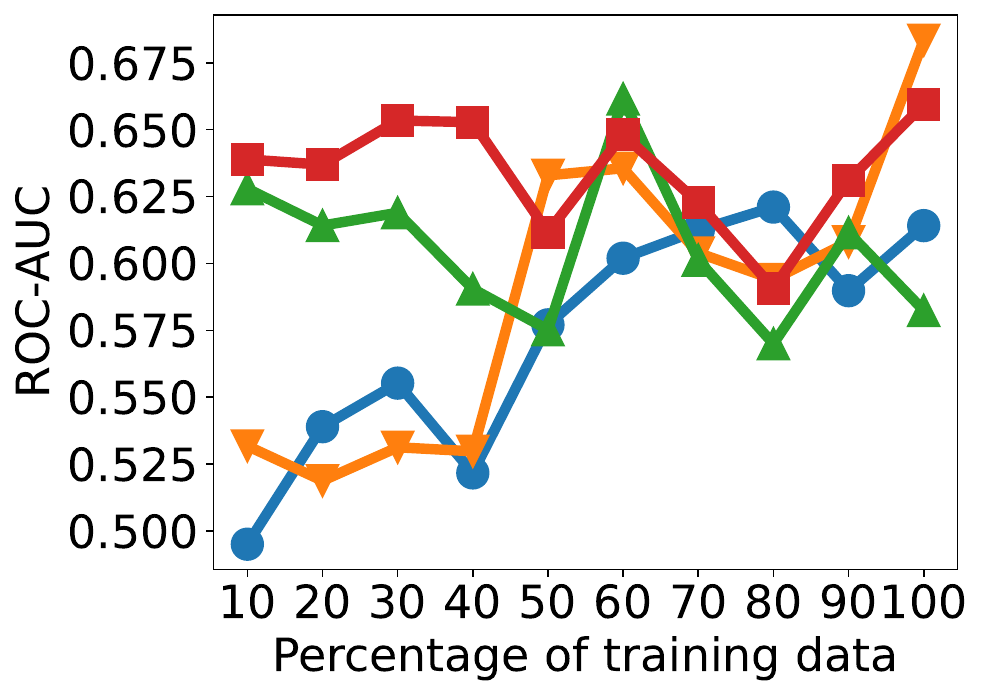}
    \vspace{-.2in}
    \caption{{CLINTOX ($\uparrow$)}}    
\end{subfigure}
\hfill
\begin{subfigure}[b]{0.195\textwidth}
    \centering
    \includegraphics[width=\textwidth]{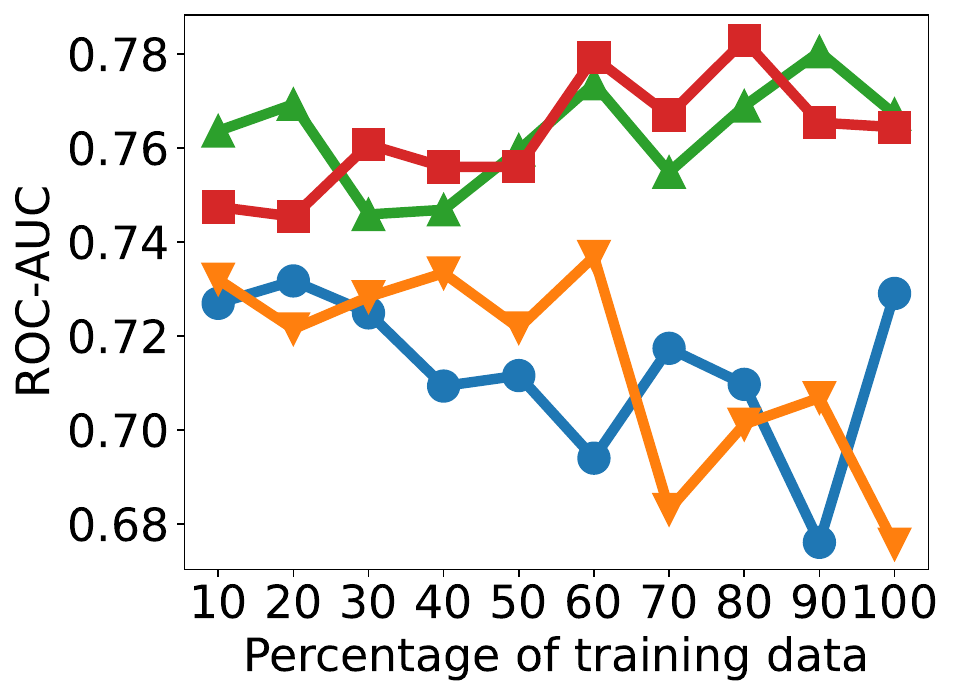}
    \vspace{-.2in}
    \caption{{HIV ($\uparrow$)}}    
\end{subfigure}
\hfill
\begin{subfigure}[b]{0.195\textwidth}
    \centering
    \includegraphics[width=\textwidth]{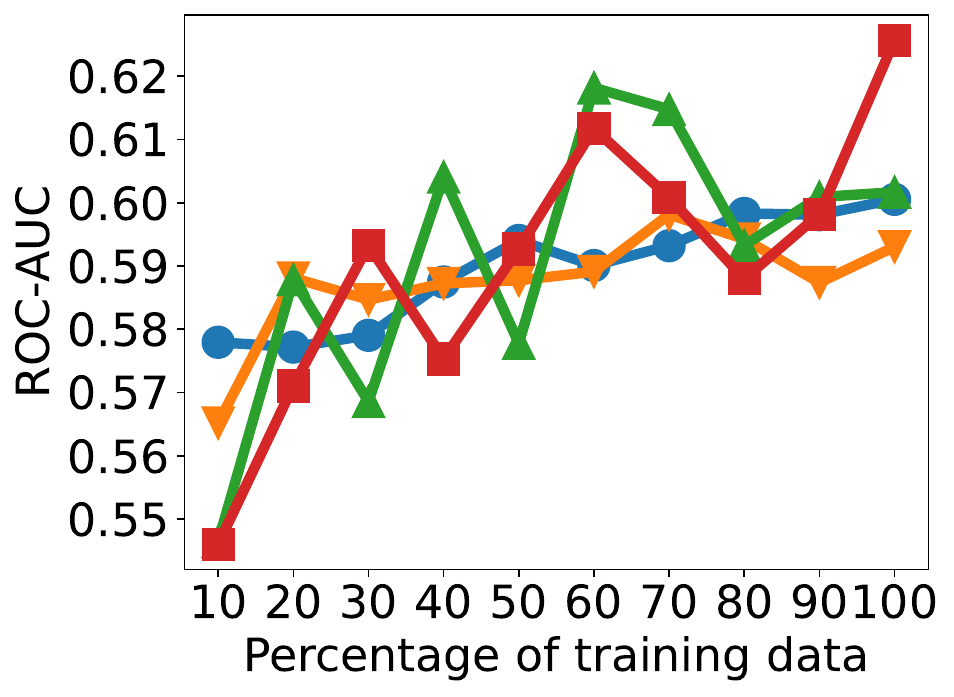}
    \vspace{-.2in}
    \caption{{SIDER ($\uparrow$)}}    
\end{subfigure}
\hfill
\begin{subfigure}[b]{0.195\textwidth}
    \centering
    \includegraphics[width=\textwidth]{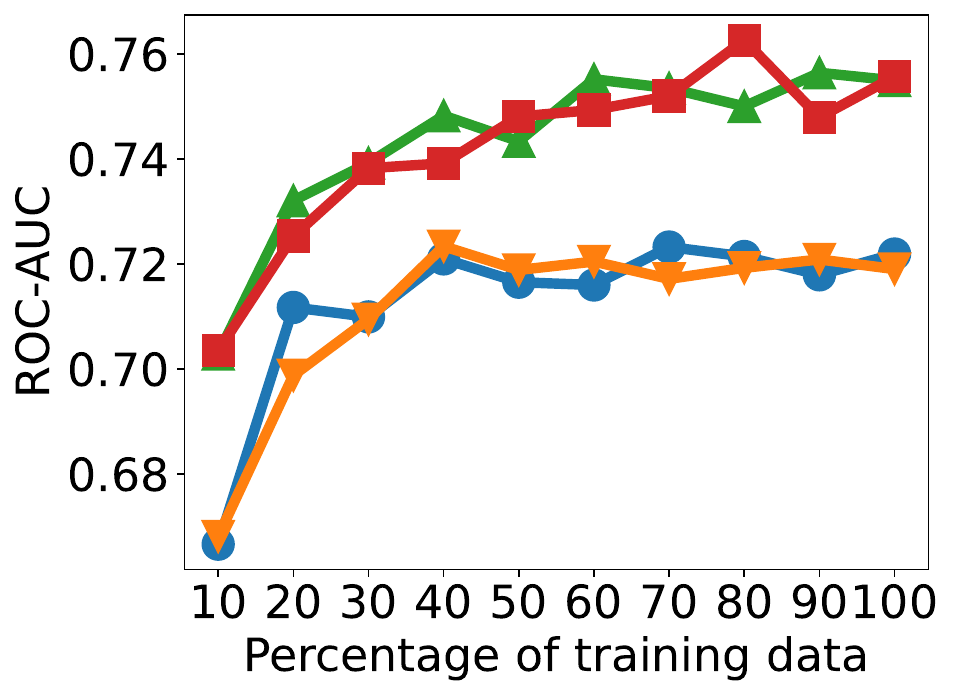}
    \vspace{-.2in}
    \caption{{TOX21 ($\uparrow$)}}    
\end{subfigure}
\hfill
\begin{subfigure}[b]{0.195\textwidth}
    \centering
    \includegraphics[width=\textwidth]{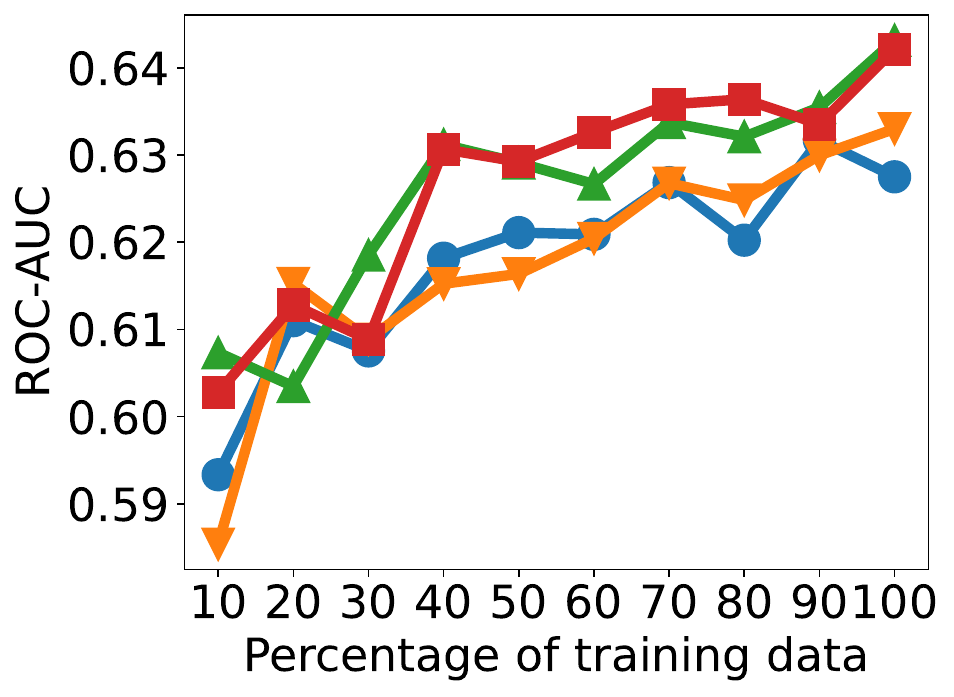}
    \vspace{-.2in}
    \caption{{TOXCAST ($\uparrow$)}}    
\end{subfigure}
\\ 
\begin{subfigure}[b]{0.195\textwidth}
    \centering
    \includegraphics[width=\textwidth]{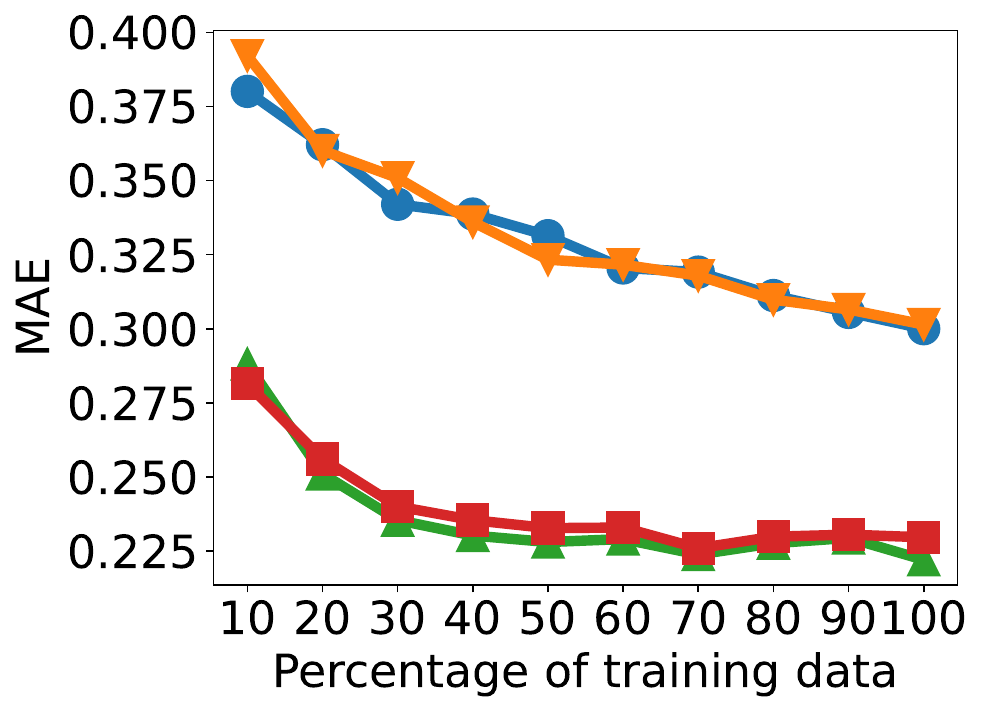}
    \vspace{-.2in}
    \caption{{BP ($\downarrow$)}}    
\end{subfigure}
\hfill
\begin{subfigure}[b]{0.195\textwidth}
    \centering
    \includegraphics[width=\textwidth]{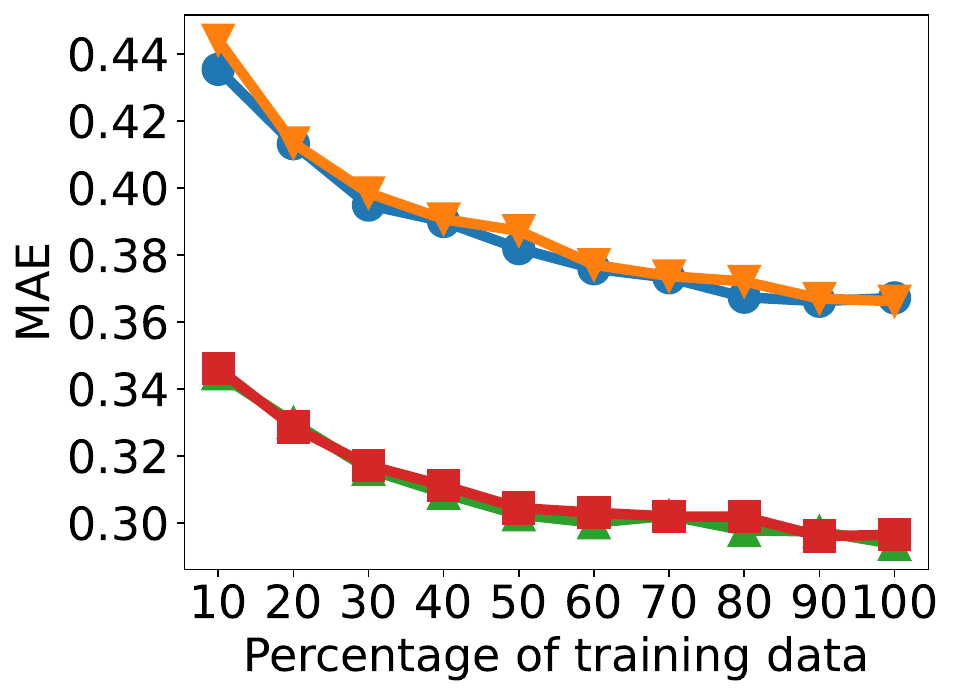}
    \vspace{-.2in}
    \caption{{MP ($\downarrow$)}}    
\end{subfigure}
\hfill
\begin{subfigure}[b]{0.195\textwidth}
    \centering
    \includegraphics[width=\textwidth]{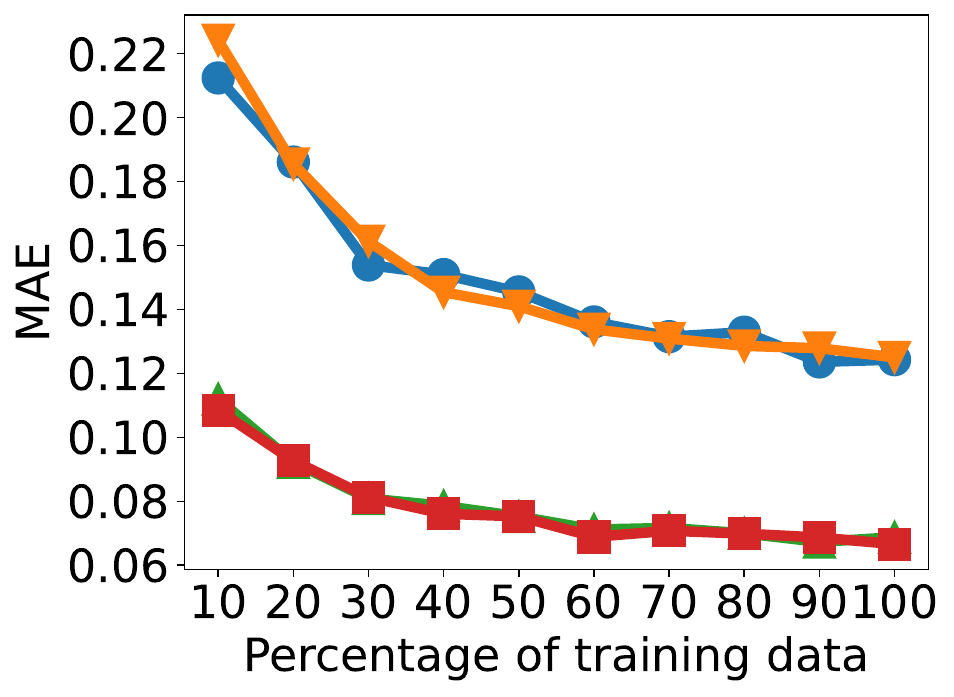}
    \vspace{-.2in}
    \caption{{RI ($\downarrow$)}}    
\end{subfigure}
\hfill
\begin{subfigure}[b]{0.195\textwidth}
    \centering
    \includegraphics[width=\textwidth]{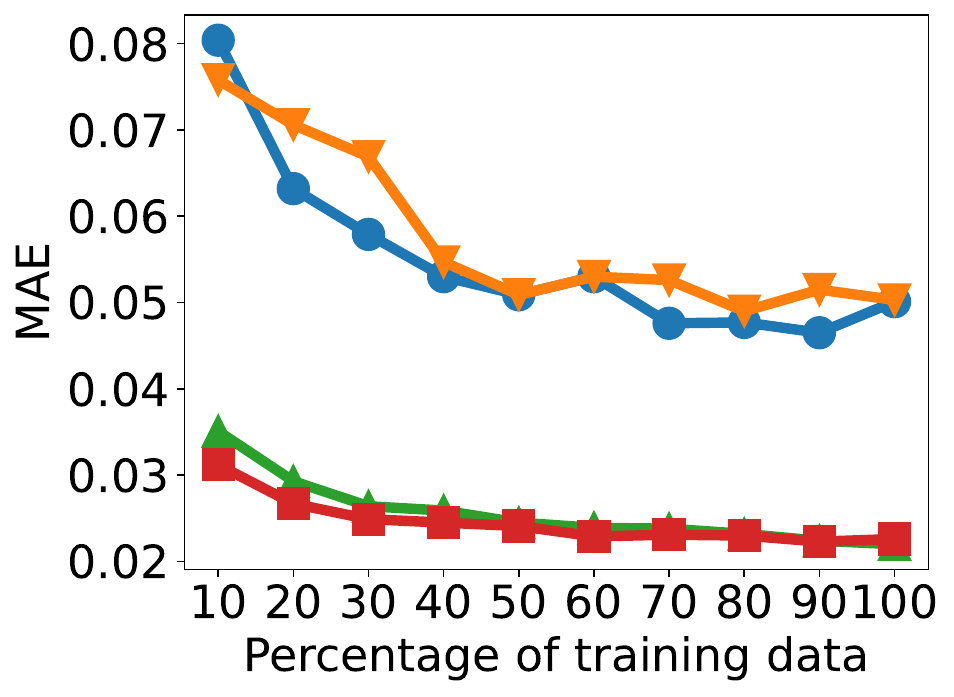}
    \vspace{-.2in}
    \caption{{logP ($\downarrow$)}}    
\end{subfigure}
\hfill
\begin{subfigure}[b]{0.195\textwidth}
    \raisebox{9mm}{\includegraphics[trim={27cm 2cm 0 2cm},clip,width=.85\textwidth,right]{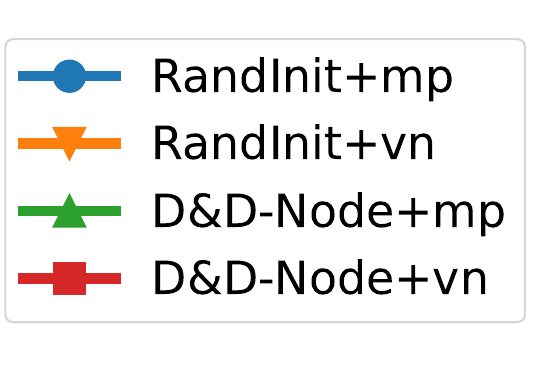}}
\end{subfigure}
\vspace{-.2in}
\caption{Experimental results demonstrating the label-efficiency of \ourmethod{} on each OGB (upper two rows) and Curated (bottom-row) task. X-axes indicate the percentage of training data used for finetuning, and Y-axes show performance measurements on the respective test sets.}
\label{fig:label_efficiency}
\vspace{-.2in}
\end{figure*}

\paragraph{\ourmethod{} transfers knowledge that is generalizable across diverse tasks.} 
Table~\ref{tab:main_results} shows finetuning results on OGB, in which each experiment is averaged across 3 runs with random seeds.
Comparing best results from \ourmethod{} and \textsc{RandInit}, our method shows superior performance in 11 out of 12 tasks, with an average performance increase of 8.8\% and 36.9\% across OGB and Curated datasets, respectively. Surprisingly, when categorizing all tasks into 3 groups (OGB-regression, OGB-classification, and Curated), the tasks on which \ourmethod{} shows the largest performance increase against \textsc{RandInit} from each group coincide with properties known to align well with 3D geometrical properties (28.1\% for LIPO, 6.58\% for HIV, and 55.99\% for logP). For instance, solving the HIV task requires assessment of the 3D shape to estimate its binding affinity with protein pockets. Both LIPO and logP tasks are tightly associated with the overall polarity of electron clouds in the molecule, which is highly associated with the spatial positioning of the atoms. This implies that denoising followed by distillation effectively transfers 3D knowledge. 



When compared against 3DInfomax, \ourmethod{} shows 15.8\% better performance for OGB datasets and 37.2\% in Curated datasets on average. This suggests that transferring molecular-specific knowledge of force fields is much more effective than contrasting representations as a proxy task; attracting and repelling molecular representations fail to fully capture generalizable similarities and discrepancies in the chemical space. Another limitation of the contrastive approach is that the gain in performance becomes limited when only a single conformer is provided per molecule~\cite{liu2021graphmvp, stark20223dinfomax}. This aligns well with our intuition that each conformer can be seen as a distribution of 3D configurations, and that learning a single local optima within the distribution does not provide much information. Meanwhile, \ourmethod{} can learn and transfer knowledge of the overall distribution with denoising and distilling, relieving practitioners from the need to obtain multiple low-energy conformers per molecule for pretraining.






\paragraph{\ourmethod{} enables label-efficient finetuning.}

To evaluate how well \ourmethod{} performs under the downstream scenario with limited number of labels, we also perform finetuning on a smaller randomly sampled subset of the original training data for each task. Figure~\ref{fig:label_efficiency} shows quantitative results of \textsc{RandInit} and \textsc{\ourmethod{}-Node}.
In 4 out of 10 OGB tasks and all 4 Curated datasets, \textsc{\ourmethod{}-Node} trained with only 10\% of training data shows accuracy comparable to that of \textsc{RandInit} trained on the full dataset. This demonstrates the utility of our approach in preventing overfitting to small data by leveraging generalizable knowledge from denoising. 
For BACE and SIDER, we find that the gain of pretraining with \ourmethod{} is relatively limited compared to other tasks. While results from other OGB tasks imply that our approach can generalize towards molecules with sizes that deviate from those in PCQM4Mv2 (average 14.1 nodes per molecule for PCQM4Mv2 vs. up to 27.0 for other OGB tasks), we conjecture that too large a difference in size (vs. 34.1 for BACE and 33.6 for SIDER) between molecules may hamper generalization.

\paragraph{\ourmethod{} is also effective on other datasets and GNN architectures.} In addition to experiments above, we also run \ourmethod{} on the QM9 benchmark~\cite{(QM9)ramakrishnan2014quantum} with 134K molecules using a different GNN architecture for the 2D student model to test whether our approach works well in a model-agnostic fashion. Specifically, we pretrain TorchMD-NET~\cite{tholke2022torchmd} on the QM9 dataset via denoising, and distill its representations from QM9 onto PNA~\cite{corso2020pna}, a message-passing GNN model previously used in~\cite{stark20223dinfomax}. We then finetune the pretrained 2D student to each of the properties in QM9. Note that we do not perform hyperparameter tuning for this task, but instead use the same default setting for PNA chosen by \cite{stark20223dinfomax} including the same set of random seeds for fair comparison. More details on the experimental setup can be found in Appendix~\ref{app:qm9}.

In Table~\ref{tab:vs_3dinfomax_qm9_short}, we observe that \ourmethod{} shows performs competitively against 3DInfomax on 7 out of 8 datasets. More interestingly, \textsc{\ourmethod{}-Node} is especially more effective than \textsc{\ourmethod{}-Graph} on QM9, significantly outperforming 3DInfomax on $\alpha$ and \textsc{gap} properties. Considering that molecules in QM9 are labeled with energy-related properties through the same DFT computation used to solve their 3D atom coordinates, we conjecture that the structured-distillation approach is more well-suited to such properties by transferring atom representations in a fine-grained manner. As we propose two variants of \ourmethod{}, it would be interesting to investigate whether our comparison between \textsc{\ourmethod{}-Node} vs. \textsc{\ourmethod{}-Graph} in empirical performance aligns with domain-specific interpretations of each chemical property, which we leave as future work.

\begin{table}[!t]
    \centering
    \caption{Average performance and standard deviations from the QM9 quantum property prediction. Baseline results are results reported by \cite{stark20223dinfomax}. Best results within one standard deviation are in \textbf{bold}.} \label{tab:vs_3dinfomax_qm9_short} 
    \vspace{.1in}
    \resizebox{\textwidth}{!}{\begin{tabular}{l|cccccccc}
        \toprule
        Target & $\mu$ & $\alpha$ & \textsc{homo} & \textsc{lumo} & \textsc{gap} & \textsc{r2} & \textsc{zpve} & $c_v$\\
        Metric & MAE($\downarrow$) & MAE($\downarrow$) & MAE($\downarrow$) & MAE($\downarrow$) & MAE($\downarrow$) & MAE($\downarrow$) & MAE($\downarrow$) & MAE($\downarrow$)\\
        \midrule
        \textsc{RandInit} & 0.4133{\scriptsize $\pm$0.003} & 0.3972{\scriptsize $\pm$0.014} & 82.10{\scriptsize $\pm$0.33} & 85.72{\scriptsize $\pm$1.62} & 123.08{\scriptsize $\pm$3.98} & 22.14{\scriptsize $\pm$0.21} & 15.08{\scriptsize $\pm$2.83} & 0.1670{\scriptsize $\pm$0.004} \\
        \textsc{3DInfomax} & \bf 0.3507{\scriptsize $\pm$0.005} & 0.3268{\scriptsize $\pm$0.006} & \bf 68.96{\scriptsize $\pm$0.32} & \bf 69.51{\scriptsize $\pm$0.54} & 101.71{\scriptsize $\pm$2.03} & \bf 17.39{\scriptsize $\pm$0.94} & \bf 7.966{\scriptsize $\pm$1.87} & 0.1306{\scriptsize $\pm$0.009} \\
        \midrule
        \textsc{D\&D-Graph} & \bf 0.3512{\scriptsize $\pm$0.005} & 0.2903{\scriptsize $\pm$0.029} & 70.36{\scriptsize $\pm$2.70} & 71.72{\scriptsize $\pm$2.17} & 98.82{\scriptsize $\pm$1.09} & \bf 17.61{\scriptsize $\pm$0.68} & 12.88{\scriptsize $\pm$3.42} & \bf 0.1248{\scriptsize $\pm$0.006}\\
        \textsc{D\&D-Node} & 0.3552{\scriptsize $\pm$0.004} & \bf 0.2807{\scriptsize $\pm$0.015} & \bf 69.32{\scriptsize $\pm$1.01} & \bf 69.63{\scriptsize $\pm$0.62} & \bf 98.79{\scriptsize $\pm$0.89} & \bf 17.75{\scriptsize $\pm$0.43} & 10.19{\scriptsize $\pm$1.77} & 0.1429{\scriptsize $\pm$0.015}\\
        \bottomrule
    \end{tabular}
    }
    \vspace{-.2in}
\end{table}

\cutsubsectionup
\section{Conclusion}\label{sec:conclusion}
\cutsubsectionup
In this paper, we propose \ourmethod{}, a novel self-supervised molecular representation learning framework that allows use of models pretrained via 3D conformer denoising towards downstream tasks that only provide 2D molecular graphs as input. As molecular force fields provide chemically generalizable information across various tasks, \ourmethod{} demonstrates significant knowledge transfer to diverse molecular property prediction tasks. Additional analyses show that \ourmethod{} is also highly label-efficient, showing significant performance boosts under downstream settings where the number of labeled training data is limited. As future work, we hope to extend \ourmethod{} towards a multitask setting~\cite{liu2020structured} to see if we can finetune a single model that performs well simultaneously across multiple molecular properties. Another exciting direction is to explore use of generative models~\cite{ganea2021geomol,xu2022geodiff,jing2022torsional} for molecular property prediction. Inspired by use of image diffusion models for semantic segmentation~\cite{baranchuk2021label}, it would be interesting to test whether intermediate representations inferred by diffusion-based generative models can be leveraged downstream towards generalizable knowledge transfer.

\bibliographystyle{plainnat}
\bibliography{neurips_2023.bib}

\begin{thebibliography}{64}
\providecommand{\natexlab}[1]{#1}
\providecommand{\url}[1]{\texttt{#1}}
\expandafter\ifx\csname urlstyle\endcsname\relax
  \providecommand{\doi}[1]{doi: #1}\else
  \providecommand{\doi}{doi: \begingroup \urlstyle{rm}\Url}\fi

\bibitem[Ba and Caruana(2014)]{ba2014deep}
Jimmy Ba and Rich Caruana.
\newblock Do deep nets really need to be deep?
\newblock \emph{Advances in neural information processing systems}, 27, 2014.

\bibitem[Bachmann et~al.(2022)Bachmann, Mizrahi, Atanov, and
  Zamir]{bachmann2022multimae}
Roman Bachmann, David Mizrahi, Andrei Atanov, and Amir Zamir.
\newblock Multimae: Multi-modal multi-task masked autoencoders.
\newblock \emph{arXiv preprint arXiv:2204.01678}, 2022.

\bibitem[Baranchuk et~al.(2021)Baranchuk, Rubachev, Voynov, Khrulkov, and
  Babenko]{baranchuk2021label}
Dmitry Baranchuk, Ivan Rubachev, Andrey Voynov, Valentin Khrulkov, and Artem
  Babenko.
\newblock Label-efficient semantic segmentation with diffusion models.
\newblock \emph{arXiv preprint arXiv:2112.03126}, 2021.

\bibitem[Bradley et~al.(2014)Bradley, Williams, and Lang]{Bradley2014}
Jean-Claude Bradley, Antony Williams, and Andrew Lang.
\newblock {Jean-Claude Bradley Open Melting Point Dataset}.
\newblock 5 2014.
\newblock \doi{10.6084/m9.figshare.1031637.v2}.
\newblock URL
  \url{https://figshare.com/articles/dataset/Jean_Claude_Bradley_Open_Melting_Point_Datset/1031637}.

\bibitem[Bronstein et~al.(2021)Bronstein, Bruna, Cohen, and
  Veli{\v{c}}kovi{\'c}]{bronstein2021geometric}
Michael~M Bronstein, Joan Bruna, Taco Cohen, and Petar Veli{\v{c}}kovi{\'c}.
\newblock Geometric deep learning: Grids, groups, graphs, geodesics, and
  gauges.
\newblock \emph{arXiv preprint arXiv:2104.13478}, 2021.

\bibitem[Brown et~al.(2020)Brown, Mann, Ryder, Subbiah, Kaplan, Dhariwal,
  Neelakantan, Shyam, Sastry, Askell, et~al.]{brown2020language}
Tom Brown, Benjamin Mann, Nick Ryder, Melanie Subbiah, Jared~D Kaplan, Prafulla
  Dhariwal, Arvind Neelakantan, Pranav Shyam, Girish Sastry, Amanda Askell,
  et~al.
\newblock Language models are few-shot learners.
\newblock \emph{Advances in neural information processing systems},
  33:\penalty0 1877--1901, 2020.

\bibitem[Bruna et~al.(2013)Bruna, Zaremba, Szlam, and
  LeCun]{https://doi.org/10.48550/arxiv.1312.6203}
Joan Bruna, Wojciech Zaremba, Arthur Szlam, and Yann LeCun.
\newblock Spectral networks and locally connected networks on graphs.
\newblock 2013.
\newblock \doi{10.48550/ARXIV.1312.6203}.
\newblock URL \url{https://arxiv.org/abs/1312.6203}.

\bibitem[Chen et~al.(2020)Chen, Kornblith, Norouzi, and Hinton]{chen2020simclr}
Ting Chen, Simon Kornblith, Mohammad Norouzi, and Geoffrey Hinton.
\newblock A simple framework for contrastive learning of visual
  representations.
\newblock In \emph{International conference on machine learning}, pages
  1597--1607. PMLR, 2020.

\bibitem[Coley et~al.(2017)Coley, Barzilay, Green, Jaakkola, and
  Jensen]{doi:10.1021/acs.jcim.6b00601}
Connor~W. Coley, Regina Barzilay, William~H. Green, Tommi~S. Jaakkola, and
  Klavs~F. Jensen.
\newblock Convolutional embedding of attributed molecular graphs for physical
  property prediction.
\newblock \emph{Journal of Chemical Information and Modeling}, 57\penalty0
  (8):\penalty0 1757--1772, 2017.
\newblock \doi{10.1021/acs.jcim.6b00601}.
\newblock URL \url{https://doi.org/10.1021/acs.jcim.6b00601}.
\newblock PMID: 28696688.

\bibitem[Corso et~al.(2020)Corso, Cavalleri, Beaini, Li{\`o}, and
  Veli{\v{c}}kovi{\'c}]{corso2020pna}
Gabriele Corso, Luca Cavalleri, Dominique Beaini, Pietro Li{\`o}, and Petar
  Veli{\v{c}}kovi{\'c}.
\newblock Principal neighbourhood aggregation for graph nets.
\newblock \emph{Advances in Neural Information Processing Systems},
  33:\penalty0 13260--13271, 2020.

\bibitem[Defferrard et~al.(2016)Defferrard, Bresson, and
  Vandergheynst]{https://doi.org/10.48550/arxiv.1606.09375}
Michaël Defferrard, Xavier Bresson, and Pierre Vandergheynst.
\newblock Convolutional neural networks on graphs with fast localized spectral
  filtering.
\newblock 2016.
\newblock \doi{10.48550/ARXIV.1606.09375}.
\newblock URL \url{https://arxiv.org/abs/1606.09375}.

\bibitem[Devlin et~al.(2018)Devlin, Chang, Lee, and Toutanova]{devlin2018bert}
Jacob Devlin, Ming-Wei Chang, Kenton Lee, and Kristina Toutanova.
\newblock Bert: Pre-training of deep bidirectional transformers for language
  understanding.
\newblock \emph{arXiv preprint arXiv:1810.04805}, 2018.

\bibitem[Duvenaud et~al.(2015)Duvenaud, Maclaurin, Aguilera-Iparraguirre,
  Gómez-Bombarelli, Hirzel, Aspuru-Guzik, and
  Adams]{https://doi.org/10.48550/arxiv.1509.09292}
David Duvenaud, Dougal Maclaurin, Jorge Aguilera-Iparraguirre, Rafael
  Gómez-Bombarelli, Timothy Hirzel, Alán Aspuru-Guzik, and Ryan~P. Adams.
\newblock Convolutional networks on graphs for learning molecular fingerprints.
\newblock 2015.
\newblock \doi{10.48550/ARXIV.1509.09292}.
\newblock URL \url{https://arxiv.org/abs/1509.09292}.

\bibitem[Fuchs et~al.(2020)Fuchs, Worrall, Fischer, and Welling]{fuchs2020se}
Fabian Fuchs, Daniel Worrall, Volker Fischer, and Max Welling.
\newblock Se (3)-transformers: 3d roto-translation equivariant attention
  networks.
\newblock \emph{Advances in Neural Information Processing Systems},
  33:\penalty0 1970--1981, 2020.

\bibitem[Ganea et~al.(2021)Ganea, Pattanaik, Coley, Barzilay, Jensen, Green,
  and Jaakkola]{ganea2021geomol}
Octavian Ganea, Lagnajit Pattanaik, Connor Coley, Regina Barzilay, Klavs
  Jensen, William Green, and Tommi Jaakkola.
\newblock Geomol: Torsional geometric generation of molecular 3d conformer
  ensembles.
\newblock \emph{Advances in Neural Information Processing Systems},
  34:\penalty0 13757--13769, 2021.

\bibitem[Gasteiger et~al.(2021)Gasteiger, Becker, and
  G{\"u}nnemann]{gasteiger2021gemnet}
Johannes Gasteiger, Florian Becker, and Stephan G{\"u}nnemann.
\newblock Gemnet: Universal directional graph neural networks for molecules.
\newblock \emph{Advances in Neural Information Processing Systems},
  34:\penalty0 6790--6802, 2021.

\bibitem[Gou et~al.(2021)Gou, Yu, Maybank, and Tao]{gou2021knowledge}
Jianping Gou, Baosheng Yu, Stephen~J Maybank, and Dacheng Tao.
\newblock Knowledge distillation: A survey.
\newblock \emph{International Journal of Computer Vision}, 129\penalty0
  (6):\penalty0 1789--1819, 2021.

\bibitem[Gupta et~al.(2016)Gupta, Hoffman, and Malik]{gupta2016cross}
Saurabh Gupta, Judy Hoffman, and Jitendra Malik.
\newblock Cross modal distillation for supervision transfer.
\newblock In \emph{Proceedings of the IEEE conference on computer vision and
  pattern recognition}, pages 2827--2836, 2016.

\bibitem[Guvench(2016)]{GUVENCH20161928}
Olgun Guvench.
\newblock Computational functional group mapping for drug discovery.
\newblock \emph{Drug Discovery Today}, 21\penalty0 (12):\penalty0 1928--1931,
  2016.
\newblock ISSN 1359-6446.
\newblock \doi{https://doi.org/10.1016/j.drudis.2016.06.030}.
\newblock URL
  \url{https://www.sciencedirect.com/science/article/pii/S1359644616302483}.

\bibitem[Hamilton(2020)]{hamilton2020graph}
W.L. Hamilton.
\newblock \emph{Graph Representation Learning}.
\newblock Synthesis lectures on artificial intelligence and machine learning.
  Morgan \& Claypool Publishers, 2020.
\newblock ISBN 9781681739632.
\newblock URL \url{https://books.google.co.kr/books?id=V6HnzQEACAAJ}.

\bibitem[Hassani and Khasahmadi(2020)]{hassani2020contrastive}
Kaveh Hassani and Amir~Hosein Khasahmadi.
\newblock Contrastive multi-view representation learning on graphs.
\newblock In \emph{International Conference on Machine Learning}, pages
  4116--4126. PMLR, 2020.

\bibitem[He et~al.(2020)He, Fan, Wu, Xie, and Girshick]{he2020momentum}
Kaiming He, Haoqi Fan, Yuxin Wu, Saining Xie, and Ross Girshick.
\newblock Momentum contrast for unsupervised visual representation learning.
\newblock In \emph{Proceedings of the IEEE/CVF conference on computer vision
  and pattern recognition}, pages 9729--9738, 2020.

\bibitem[Hinton et~al.(2015)Hinton, Vinyals, Dean,
  et~al.]{hinton2015distilling}
Geoffrey Hinton, Oriol Vinyals, Jeff Dean, et~al.
\newblock Distilling the knowledge in a neural network.
\newblock \emph{arXiv preprint arXiv:1503.02531}, 2\penalty0 (7), 2015.

\bibitem[Hou et~al.(2022)Hou, Liu, Cen, Dong, Yang, Wang, and
  Tang]{(graphmae)hou2022graphmae}
Zhenyu Hou, Xiao Liu, Yukuo Cen, Yuxiao Dong, Hongxia Yang, Chunjie Wang, and
  Jie Tang.
\newblock Graphmae: Self-supervised masked graph autoencoders.
\newblock In \emph{Proceedings of the 28th ACM SIGKDD Conference on Knowledge
  Discovery and Data Mining}, pages 594--604, 2022.

\bibitem[Hu et~al.(2019)Hu, Liu, Gomes, Zitnik, Liang, Pande, and
  Leskovec]{hu2019strategies}
Weihua Hu, Bowen Liu, Joseph Gomes, Marinka Zitnik, Percy Liang, Vijay Pande,
  and Jure Leskovec.
\newblock Strategies for pre-training graph neural networks.
\newblock \emph{arXiv preprint arXiv:1905.12265}, 2019.

\bibitem[Hu et~al.(2020)Hu, Fey, Zitnik, Dong, Ren, Liu, Catasta, and
  Leskovec]{hu2020ogb}
Weihua Hu, Matthias Fey, Marinka Zitnik, Yuxiao Dong, Hongyu Ren, Bowen Liu,
  Michele Catasta, and Jure Leskovec.
\newblock Open graph benchmark: Datasets for machine learning on graphs.
\newblock \emph{Advances in neural information processing systems},
  33:\penalty0 22118--22133, 2020.

\bibitem[Jin et~al.(2018)Jin, Yang, Barzilay, and
  Jaakkola]{https://doi.org/10.48550/arxiv.1812.01070}
Wengong Jin, Kevin Yang, Regina Barzilay, and Tommi Jaakkola.
\newblock Learning multimodal graph-to-graph translation for molecular
  optimization, 2018.
\newblock URL \url{https://arxiv.org/abs/1812.01070}.

\bibitem[Jing et~al.(2022)Jing, Corso, Chang, Barzilay, and
  Jaakkola]{jing2022torsional}
Bowen Jing, Gabriele Corso, Jeffrey Chang, Regina Barzilay, and Tommi Jaakkola.
\newblock Torsional diffusion for molecular conformer generation.
\newblock \emph{arXiv preprint arXiv:2206.01729}, 2022.

\bibitem[Kanakaveti et~al.(2017)Kanakaveti, Sakthivel, Rayala, and
  Gromiha]{https://doi.org/10.1111/cbdd.12952}
Vishnupriya Kanakaveti, Ramasamy Sakthivel, S.~K. Rayala, and M.~Michael
  Gromiha.
\newblock Importance of functional groups in predicting the activity of small
  molecule inhibitors for bcl-2 and bcl-xl.
\newblock \emph{Chemical Biology \& Drug Design}, 90\penalty0 (2):\penalty0
  308--316, 2017.
\newblock \doi{https://doi.org/10.1111/cbdd.12952}.
\newblock URL \url{https://onlinelibrary.wiley.com/doi/abs/10.1111/cbdd.12952}.

\bibitem[Kim et~al.(2022{\natexlab{a}})Kim, Baek, and
  Hwang]{(D-SLA)kim2022graph}
Dongki Kim, Jinheon Baek, and Sung~Ju Hwang.
\newblock Graph self-supervised learning with accurate discrepancy learning.
\newblock \emph{arXiv preprint arXiv:2202.02989}, 2022{\natexlab{a}}.

\bibitem[Kim et~al.(2022{\natexlab{b}})Kim, Nguyen, Min, Cho, Lee, Lee, and
  Hong]{kim2022tokengt}
Jinwoo Kim, Tien~Dat Nguyen, Seonwoo Min, Sungjun Cho, Moontae Lee, Honglak
  Lee, and Seunghoon Hong.
\newblock Pure transformers are powerful graph learners.
\newblock \emph{arXiv preprint arXiv:2207.02505}, 2022{\natexlab{b}}.

\bibitem[Kim et~al.(2022{\natexlab{c}})Kim, Chen, Cheng, Gindulyte, He, He, Li,
  Shoemaker, Thiessen, Yu, Zaslavsky, Zhang, and Bolton]{10.1093/nar/gkac956}
Sunghwan Kim, Jie Chen, Tiejun Cheng, Asta Gindulyte, Jia He, Siqian He,
  Qingliang Li, Benjamin~A Shoemaker, Paul~A Thiessen, Bo~Yu, Leonid Zaslavsky,
  Jian Zhang, and Evan~E Bolton.
\newblock {PubChem 2023 update}.
\newblock \emph{Nucleic Acids Research}, 51\penalty0 (D1):\penalty0
  D1373--D1380, 10 2022{\natexlab{c}}.
\newblock ISSN 0305-1048.
\newblock \doi{10.1093/nar/gkac956}.
\newblock URL \url{https://doi.org/10.1093/nar/gkac956}.

\bibitem[Landrum(2016)]{landrum2016rdkit}
Greg Landrum.
\newblock Rdkit: Open-source cheminformatics software.
\newblock 2016.
\newblock URL
  \url{https://github.com/rdkit/rdkit/releases/tag/Release_2016_09_4}.

\bibitem[Li et~al.(2022)Li, Zhao, and Zeng]{(KPGT)li2022kpgt}
Han Li, Dan Zhao, and Jianyang Zeng.
\newblock Kpgt: Knowledge-guided pre-training of graph transformer for
  molecular property prediction.
\newblock \emph{arXiv preprint arXiv:2206.03364}, 2022.

\bibitem[Liu et~al.(2021)Liu, Wang, Liu, Lasenby, Guo, and
  Tang]{liu2021graphmvp}
Shengchao Liu, Hanchen Wang, Weiyang Liu, Joan Lasenby, Hongyu Guo, and Jian
  Tang.
\newblock Pre-training molecular graph representation with 3d geometry.
\newblock \emph{arXiv preprint arXiv:2110.07728}, 2021.

\bibitem[Liu et~al.(2022{\natexlab{a}})Liu, Guo, and Tang]{liu2022molecular}
Shengchao Liu, Hongyu Guo, and Jian Tang.
\newblock Molecular geometry pretraining with se (3)-invariant denoising
  distance matching.
\newblock \emph{arXiv preprint arXiv:2206.13602}, 2022{\natexlab{a}}.

\bibitem[Liu et~al.(2022{\natexlab{b}})Liu, Qu, Zhang, Cai, and
  Tang]{liu2022structured}
Shengchao Liu, Meng Qu, Zuobai Zhang, Huiyu Cai, and Jian Tang.
\newblock Structured multi-task learning for molecular property prediction.
\newblock In \emph{International Conference on Artificial Intelligence and
  Statistics}, pages 8906--8920. PMLR, 2022{\natexlab{b}}.

\bibitem[Liu et~al.(2020)Liu, Shu, Wang, and Shen]{liu2020structured}
Yifan Liu, Changyong Shu, Jingdong Wang, and Chunhua Shen.
\newblock Structured knowledge distillation for dense prediction.
\newblock \emph{IEEE transactions on pattern analysis and machine
  intelligence}, 2020.

\bibitem[Mezey(2001)]{mezey2001distributions}
Paul~G. Mezey.
\newblock Distributions and averages of molecular conformations.
\newblock \emph{Computers \& Chemistry}, 25\penalty0 (1):\penalty0 69--75,
  2001.
\newblock ISSN 0097-8485.
\newblock \doi{https://doi.org/10.1016/S0097-8485(00)00089-9}.
\newblock URL
  \url{https://www.sciencedirect.com/science/article/pii/S0097848500000899}.

\bibitem[Nakata and Shimazaki(2017)]{nakata2017pcqm4mv2}
Maho Nakata and Tomomi Shimazaki.
\newblock Pubchemqc project: A large-scale first-principles electronic
  structure database for data-driven chemistry.
\newblock \emph{Journal of Chemical Information and Modeling}, 57\penalty0
  (6):\penalty0 1300--1308, 2017.
\newblock \doi{10.1021/acs.jcim.7b00083}.
\newblock URL \url{https://doi.org/10.1021/acs.jcim.7b00083}.
\newblock PMID: 28481528.

\bibitem[Parr and Weitao(1995)]{parr1995dft}
Robert~G Parr and Yang Weitao.
\newblock \emph{{Density-Functional Theory of Atoms and Molecules}}.
\newblock Oxford University Press, 01 1995.
\newblock ISBN 9780195092769.
\newblock \doi{10.1093/oso/9780195092769.001.0001}.
\newblock URL \url{https://doi.org/10.1093/oso/9780195092769.001.0001}.

\bibitem[Pyzer-Knapp et~al.(2015)Pyzer-Knapp, Li, and
  Aspuru-Guzik]{https://doi.org/10.1002/adfm.201501919}
Edward~O. Pyzer-Knapp, Kewei Li, and Alan Aspuru-Guzik.
\newblock Learning from the harvard clean energy project: The use of neural
  networks to accelerate materials discovery.
\newblock \emph{Advanced Functional Materials}, 25\penalty0 (41):\penalty0
  6495--6502, 2015.
\newblock \doi{https://doi.org/10.1002/adfm.201501919}.
\newblock URL
  \url{https://onlinelibrary.wiley.com/doi/abs/10.1002/adfm.201501919}.

\bibitem[{Pyzer-Knapp} et~al.(2022){Pyzer-Knapp}, {Pitera}, {Staar}, {Takeda},
  {Laino}, {Sanders}, {Sexton}, {Smith}, and {Curioni}]{2022npjCM...8...84P}
Edward~O. {Pyzer-Knapp}, Jed~W. {Pitera}, Peter W.~J. {Staar}, Seiji {Takeda},
  Teodoro {Laino}, Daniel~P. {Sanders}, James {Sexton}, John~R. {Smith}, and
  Alessandro {Curioni}.
\newblock {Accelerating materials discovery using artificial intelligence, high
  performance computing and robotics}.
\newblock \emph{npj Computational Mathematics}, 8:\penalty0 84, January 2022.
\newblock \doi{10.1038/s41524-022-00765-z}.

\bibitem[Ramakrishnan et~al.(2014)Ramakrishnan, Dral, Rupp, and
  Von~Lilienfeld]{(QM9)ramakrishnan2014quantum}
Raghunathan Ramakrishnan, Pavlo~O Dral, Matthias Rupp, and O~Anatole
  Von~Lilienfeld.
\newblock Quantum chemistry structures and properties of 134 kilo molecules.
\newblock \emph{Scientific data}, 1\penalty0 (1):\penalty0 1--7, 2014.

\bibitem[Riniker and Landrum(2015)]{riniker2015etkdg}
Sereina Riniker and Gregory~A. Landrum.
\newblock Better informed distance geometry: Using what we know to improve
  conformation generation.
\newblock \emph{Journal of Chemical Information and Modeling}, 55\penalty0
  (12):\penalty0 2562--2574, 2015.
\newblock \doi{10.1021/acs.jcim.5b00654}.
\newblock URL \url{https://doi.org/10.1021/acs.jcim.5b00654}.
\newblock PMID: 26575315.

\bibitem[Rong et~al.(2020)Rong, Bian, Xu, Xie, Wei, Huang, and
  Huang]{rong2020self}
Yu~Rong, Yatao Bian, Tingyang Xu, Weiyang Xie, Ying Wei, Wenbing Huang, and
  Junzhou Huang.
\newblock Self-supervised graph transformer on large-scale molecular data.
\newblock \emph{Advances in Neural Information Processing Systems},
  33:\penalty0 12559--12571, 2020.

\bibitem[Rothermel et~al.(2021)Rothermel, Li, Rockt{\"a}schel, and
  Foerster]{rothermel2021don}
Danielle Rothermel, Margaret Li, Tim Rockt{\"a}schel, and Jakob Foerster.
\newblock Don't sweep your learning rate under the rug: A closer look at
  cross-modal transfer of pretrained transformers.
\newblock \emph{arXiv preprint arXiv:2107.12460}, 2021.

\bibitem[Satorras et~al.(2021)Satorras, Hoogeboom, and
  Welling]{satorras2021egnn}
V{\i}ctor~Garcia Satorras, Emiel Hoogeboom, and Max Welling.
\newblock E(n) equivariant graph neural networks.
\newblock In \emph{International conference on machine learning}, pages
  9323--9332. PMLR, 2021.

\bibitem[Scarselli et~al.(2009)Scarselli, Gori, Tsoi, Hagenbuchner, and
  Monfardini]{4700287}
Franco Scarselli, Marco Gori, Ah~Chung Tsoi, Markus Hagenbuchner, and Gabriele
  Monfardini.
\newblock The graph neural network model.
\newblock \emph{IEEE Transactions on Neural Networks}, 20\penalty0
  (1):\penalty0 61--80, 2009.
\newblock \doi{10.1109/TNN.2008.2005605}.

\bibitem[{Schmidt} et~al.(2019){Schmidt}, {Marques}, {Botti}, and
  {Marques}]{2019npjCM...5...83S}
Jonathan {Schmidt}, M{\'a}rio R.~G. {Marques}, Silvana {Botti}, and Miguel
  A.~L. {Marques}.
\newblock {Recent advances and applications of machine learning in solid-state
  materials science}.
\newblock \emph{npj Computational Mathematics}, 5:\penalty0 83, August 2019.
\newblock \doi{10.1038/s41524-019-0221-0}.

\bibitem[St{\"a}rk et~al.(2022)St{\"a}rk, Beaini, Corso, Tossou, Dallago,
  G{\"u}nnemann, and Li{\`o}]{stark20223dinfomax}
Hannes St{\"a}rk, Dominique Beaini, Gabriele Corso, Prudencio Tossou, Christian
  Dallago, Stephan G{\"u}nnemann, and Pietro Li{\`o}.
\newblock 3d infomax improves gnns for molecular property prediction.
\newblock In \emph{International Conference on Machine Learning}, pages
  20479--20502. PMLR, 2022.

\bibitem[Suh et~al.(2020)Suh, Fare, Warren, and
  Pyzer-Knapp]{doi:10.1146/annurev-matsci-082019-105100}
Changwon Suh, Clyde Fare, James~A. Warren, and Edward~O. Pyzer-Knapp.
\newblock Evolving the materials genome: How machine learning is fueling the
  next generation of materials discovery.
\newblock \emph{Annual Review of Materials Research}, 50\penalty0 (1):\penalty0
  1--25, 2020.
\newblock \doi{10.1146/annurev-matsci-082019-105100}.
\newblock URL \url{https://doi.org/10.1146/annurev-matsci-082019-105100}.

\bibitem[Sun et~al.(2021)Sun, Xing, Wang, Chen, and Zhou]{(MoCL)sun2021mocl}
Mengying Sun, Jing Xing, Huijun Wang, Bin Chen, and Jiayu Zhou.
\newblock Mocl: Data-driven molecular fingerprint via knowledge-aware
  contrastive learning from molecular graph.
\newblock In \emph{Proceedings of the 27th ACM SIGKDD Conference on Knowledge
  Discovery \& Data Mining}, pages 3585--3594, 2021.

\bibitem[Sun(2022)]{sun2022does}
Ruoxi Sun.
\newblock Does gnn pretraining help molecular representation?
\newblock \emph{arXiv preprint arXiv:2207.06010}, 2022.

\bibitem[Th{\"o}lke and De~Fabritiis(2022)]{tholke2022torchmd}
Philipp Th{\"o}lke and Gianni De~Fabritiis.
\newblock Torchmd-net: Equivariant transformers for neural network based
  molecular potentials.
\newblock \emph{arXiv preprint arXiv:2202.02541}, 2022.

\bibitem[Tian et~al.(2019)Tian, Krishnan, and Isola]{tian2019contrastive}
Yonglong Tian, Dilip Krishnan, and Phillip Isola.
\newblock Contrastive representation distillation.
\newblock \emph{arXiv preprint arXiv:1910.10699}, 2019.

\bibitem[Trivedi et~al.(2022)Trivedi, Lubana, Heimann, Koutra, and
  Thiagarajan]{trivedi2022analyzing}
Puja Trivedi, Ekdeep~Singh Lubana, Mark Heimann, Danai Koutra, and Jayaraman~J
  Thiagarajan.
\newblock Analyzing data-centric properties for contrastive learning on graphs.
\newblock \emph{arXiv preprint arXiv:2208.02810}, 2022.

\bibitem[Xia et~al.(2022)Xia, Wu, Chen, Hu, and Li]{(SimGRACE)xia2022simgrace}
Jun Xia, Lirong Wu, Jintao Chen, Bozhen Hu, and Stan~Z Li.
\newblock Simgrace: A simple framework for graph contrastive learning without
  data augmentation.
\newblock In \emph{Proceedings of the ACM Web Conference 2022}, pages
  1070--1079, 2022.

\bibitem[Xu et~al.(2022)Xu, Yu, Song, Shi, Ermon, and Tang]{xu2022geodiff}
Minkai Xu, Lantao Yu, Yang Song, Chence Shi, Stefano Ermon, and Jian Tang.
\newblock Geodiff: A geometric diffusion model for molecular conformation
  generation.
\newblock \emph{arXiv preprint arXiv:2203.02923}, 2022.

\bibitem[Yaws(2015)]{YAWS20151}
Carl~L. Yaws.
\newblock Chapter 1 - physical properties – organic compounds.
\newblock In Carl~L. Yaws, editor, \emph{The Yaws Handbook of Physical
  Properties for Hydrocarbons and Chemicals (Second Edition)}, pages 1--683.
  Gulf Professional Publishing, Boston, second edition edition, 2015.
\newblock ISBN 978-0-12-800834-8.
\newblock \doi{https://doi.org/10.1016/B978-0-12-800834-8.00001-3}.
\newblock URL
  \url{https://www.sciencedirect.com/science/article/pii/B9780128008348000013}.

\bibitem[You et~al.(2020)You, Chen, Sui, Chen, Wang, and Shen]{you2020graphcl}
Yuning You, Tianlong Chen, Yongduo Sui, Ting Chen, Zhangyang Wang, and Yang
  Shen.
\newblock Graph contrastive learning with augmentations.
\newblock \emph{Advances in Neural Information Processing Systems},
  33:\penalty0 5812--5823, 2020.

\bibitem[Zaidi et~al.(2022)Zaidi, Schaarschmidt, Martens, Kim, Teh,
  Sanchez-Gonzalez, Battaglia, Pascanu, and Godwin]{zaidi2022pvd}
Sheheryar Zaidi, Michael Schaarschmidt, James Martens, Hyunjik Kim, Yee~Whye
  Teh, Alvaro Sanchez-Gonzalez, Peter Battaglia, Razvan Pascanu, and Jonathan
  Godwin.
\newblock Pre-training via denoising for molecular property prediction.
\newblock \emph{arXiv preprint arXiv:2206.00133}, 2022.

\bibitem[Zhang et~al.(2021)Zhang, Liu, Wang, Lu, and
  Lee]{(MGSSL)zhang2021motif}
Zaixi Zhang, Qi~Liu, Hao Wang, Chengqiang Lu, and Chee-Kong Lee.
\newblock Motif-based graph self-supervised learning for molecular property
  prediction.
\newblock \emph{Advances in Neural Information Processing Systems},
  34:\penalty0 15870--15882, 2021.

\bibitem[Zhu et~al.(2022)Zhu, Xia, Wu, Xie, Qin, Zhou, Li, and
  Liu]{(unimol)zhu2022unified}
Jinhua Zhu, Yingce Xia, Lijun Wu, Shufang Xie, Tao Qin, Wengang Zhou, Houqiang
  Li, and Tie-Yan Liu.
\newblock Unified 2d and 3d pre-training of molecular representations.
\newblock In \emph{Proceedings of the 28th ACM SIGKDD Conference on Knowledge
  Discovery and Data Mining}, pages 2626--2636, 2022.

\end{thebibliography}

\newpage
\appendix
\appendix
\onecolumn

\newcounter{questioncntr}
\newcommand*{\question}{%
    \stepcounter{questioncntr}%
    \textbf{Q\arabic{questioncntr}: }%
}
\newcommand*{\answer}{%
    \textbf{A\arabic{questioncntr}: }%
}

\setcounter{section}{16}
\section*{Answers to Potential Questions}
\cutsubsectiondown

To provide a better understanding of our draft, we start the supplementary material by providing answers to potential questions on the overall motivation of our work, our proposed methodology, and presented empirical results. We hope most questions during review can be answered in this section, and would be happy to clarify any further questions during the author response period as well. Further supplementary material can be found in the following sections.

\question \textbf{What is novel about \ourmethod{}, considering that it combines pretraining via denoising~\cite{zaidi2022pvd} and cross-modal distillation~\cite{gupta2016cross} from previous work?} \\
\answer While the two components \ourmethod{} are indeed from separate existing work, our work is the first to combine the two techniques for molecular property prediction and show that denoising 3D conformers can be used independently as a powerful pretraining objective that transfers generalizable knowledge to 2D graph encoders for molecular property prediction. Unlike previous methods, \ourmethod{} allows 2D graph encoders to learn force fields on the 3D space, the information from which we conjecture to be critical in molecular property prediction. Experiments with OGB and QM9 testbeds show that our approach is effective despite its simplicity.

\question \textbf{Why is using one conformer per molecule better? Why not use more conformers for \ourmethod{}?}\\
\answer Using less conformers is better in terms of efficiency since each conformer requires expensive optimizations via DFT. Therefore, we test \ourmethod{} on a setting where each molecule only has one conformer. Nonetheless, \ourmethod{} can still be used with multiple conformers per molecule, in which case we expect to see performance improvements as denoising from multiple local optima can give a broader information on the force field of the molecule.

\question \textbf{What are the node and edge features used for molecular graphs?}\\
\answer We use the featurization pipeline provided by the OGB library~\cite{hu2020ogb}. For reference, below is a list of node- and edge-features used in our experiments:
\vspace{-5px}
\begin{itemize}
    \item 9 node features: atomic number, chirality, degree, charge, number of hydrogens attached, number of radical electrons, hybridization, aromaticity, in a ring.
    \item 3 edge features: bond type, bond stereo configuration, conjugation
\end{itemize}

\question \textbf{\ourmethod{} should be evaluated using more baselines and architectures.} \\
\answer We are aware of many other baselines such as GraphMVP~\cite{liu2021graphmvp}, GraphMAE~\cite{(graphmae)hou2022graphmae}, MGSSL~\cite{(MGSSL)zhang2021motif}, SimGRACE~\cite{(SimGRACE)xia2022simgrace}, MoCL~\cite{(MoCL)sun2021mocl}, D-SLA~\cite{(D-SLA)kim2022graph}, UnifiedMol~\cite{(unimol)zhu2022unified}, and KPGT~\cite{(KPGT)li2022kpgt}. For this paper, however, we chose 3DInfomax~\cite{stark20223dinfomax} as a representative baseline due to it being the first to propose a molecular representation learning pipeline that uses 3D conformers only during pretraining. While we chose to mainly test \ourmethod{} using a global-attention based architecture (\textit{i.e.} TokenGT), integrating its implementation with other pretraining techniques is a significant bottleneck at this time, but we hope to run further experiments soon. We also hope to test \ourmethod{} with message-passing GNNs (\textit{i.e.} GCN, GIN, GraphSAGE), but for this work, we focus on training with TokenGT\cite{kim2022tokengt} due to its strong theoretical expressiveness that allows distilling 3D conformer representations as closely as possible.

\question \textbf{Why does \ourmethod{} use L1- and BCE-loss for regression and classification for finetuning?} \\
\answer These choices of loss functions are from previous work~\cite{stark20223dinfomax}. While it is possible to add additional loss terms for \ourmethod{} such as regularization, we follow the same choices for fair comparison. Furthermore, we do not use additional loss terms to directly examine the overall quality of pretrained representations from the 3D denoiser.

\question \textbf{How much overlap is there between each curated dataset vs. PCQM4Mv2?}\\
\answer In Table~\ref{tab:curated_overlap}, we provide data overlap measurements of our curated datasets against the pretraining PCQM4Mv2 dataset in 3 different aspects: \textit{Elements} shows the percentage of elements in each dataset that also appears in PCQM4Mv2. \textit{Composition} shows a similar percentage in molecular composition instead, which is computed by the number of each element present in the molecule (excluding hydrogen atoms). Lastly, \textit{Molecule} measures the percentage of full molecules that are also in PCQM4Mv2 (matched via canonical SMILES). In terms of element converage, we find that 50\% of elements in BP are not observed during pretraining on PCQM4Mv2, while other datasets show larger overlap. The compositional overlaps are overall much higher, which implies that the molecular sizes are relatively similar. Lastly, each curated dataset shows an overlap of 10\% to 20\% in actual molecules, with logP showing the least overlap of 6.32\%.

\begin{table}[!h]
    \centering
    \caption{Overlap between 4 curated datasets and PCQM4Mv2.} \label{tab:curated_overlap} 
    \vspace{.1in}
    \resizebox{.5\textwidth}{!}{\begin{tabular}{l|cccc}
        \toprule
        Dataset & BP & MP & RI & logP\\
        \midrule
        Elements & 50.0\% & 65.2\% & 100.0\% & 68.2\% \\
        Composition & 95.0\% & 79.2\% & 96.1\% & 81.8\% \\
        Molecule & 22.4\% & 11.5\% & 18.1\% & 6.32\% \\
        \bottomrule
    \end{tabular}
    }
\end{table}

\question \textbf{Performances fluctuate between \textsc{\ourmethod{}-Graph} and \textsc{\ourmethod{}-Node}. What would be the preferred approach given a new target?}\\
\answer Unfortunately, we do not have an intuitive way to choose the better setting for distillation and thus would need to pick through hyperparameter tuning, but we conjecture that the preferred approach may be aligned with chemical characteristics of the target molecular property. This is supported by our QM9 experiments which show that \textsc{\ourmethod{}-Node} is overall superior to \textsc{\ourmethod{}-Graph} for quantum properties, which aligns with previous observations where accurate atom-wise coordinates are needed for high-quality prediction for such targets~\cite{stark20223dinfomax}. A deeper understanding of domain-knowledge would allow us to make similar analogies for physical chemistry, physiology, or biophysics targets in OGB, but for now we leave such further analysis as future work.

\setcounter{section}{0}

\cutsectiondown
\section{Dataset Statistics}\label{app:dataset}
\cutsectiondown

In Table~\ref{tab:datasets}, we provide basic statistics of each pretraining and finetuning dataset.

\begin{table}[!ht]
    \centering
    \caption{Statistics of every dataset used in our experiments.} \label{tab:datasets} 
    \vspace{.1in}
    \resizebox{\textwidth}{!}{\begin{tabular}{l|cccccc}
        \toprule
        Dataset & \# of molecules & Avg. \# of Atoms & Avg. \# of Bonds & \# of Elements & Task (Metric) & \# of Targets\\
        \midrule
        PCQM4Mv2 & 3,378,309 & 14.1 & 14.6 & 19 & - & -\\
        \midrule
        ESOL & 1,128 & 13.3 & 13.7 & 9 & \multirow{3}*{Regression (RMSE)} & 1\\
        LIPO & 4,200 & 27.0 & 29.5 & 12 & & 1\\
        FREESOLV & 642 & 8.7 & 8.4 & 9 & & 1\\
        \midrule
        BACE & 1,513 & 34.1 & 36.9 & 8 & \multirow{7}*{Binary Classification (ROC-AUC)} & 1\\
        BBBP & 2,039 & 24.1 & 26.0 & 13 & & 1\\
        CLINTOX & 1,477 & 26.2 & 27.9 & 28 & & 2\\
        HIV & 41,127 & 25.5 & 27.5 & 55 & & 1\\
        SIDER & 1,427 & 33.6 & 35.4 & 40 & & 27\\
        TOX21 & 7,831 & 18.6 & 19.3 & 51 & & 12\\
        TOXCAST & 8,576 & 18.8 & 19.3 & 53 & & 617\\
        \midrule
        BP & 7,111 & 11.1 & 10.8 & 28 & \multirow{5}*{Regression (MAE)} & 1\\
        MP & 21,744 & 15.7 & 16.4 & 22 & & 1\\
        RI & 7,304 & 12.6 & 12.5 & 6 & & 1\\
        logP & 29.458 & 40.5 & 40.5 & 19 & & 1\\
        QM9 & 130,831 & 18.0 & 37.3 & 5 & & 8 \\
        \bottomrule
    \end{tabular}
    }
\end{table}

\cutsectiondown
\section{Hyperparameters}\label{app:hyperparam}
\cutsectiondown

Table~\ref{tab:model_hyperparam} shows the hyperparameters used for our experiments.

\begin{table}[!h]
    \centering 
    \caption{Hyperparameter settings used (a,b) for the 3D teacher and 2D student architectures, (c) training parameters used for all 3 steps of \ourmethod{}.}\label{tab:model_hyperparam}
    \begin{minipage}[t]{.32\textwidth}
        \centering  
        \begin{subtable}[b]{.9\textwidth}
            \caption{3D Teacher TorchMD-Net}
            \centering 
            \resizebox{\textwidth}{!}{
            \begin{tabular}[t]{cc}
                \toprule
                 Parameter & Value \\
                \midrule
                \# of layers & 8 \\
                \# of heads & 8\\
                Embed. Dim & 768\\
                \# of rbf kernels & 64\\
                Batch size & 70\\
                \# of epochs & 30\\
                LR & 1e-7\\
                LR Scheduler & Cosine\\
                Warmup steps & 10000\\
                \bottomrule
            \end{tabular}
            }
        \end{subtable}
    \end{minipage}
    \hfill
    \begin{minipage}[t]{.32\textwidth}
        \centering  
        \begin{subtable}[b]{.85\textwidth}
            \caption{2D Student TokenGT}
            \centering 
            \resizebox{\textwidth}{!}{
            \begin{tabular}[t]{cc}
                \toprule
                 Parameter & Value \\
                \midrule
                \# of layers & 12 \\  
                \# of heads & 32 \\
                Embed. Dim. & 128 \\
                \bottomrule
            \end{tabular}
            }
        \end{subtable}
    \end{minipage}
    \hfill
    \begin{minipage}[t]{.32\textwidth}
    \centering  
        \begin{subtable}[b]{\textwidth}
            \caption{Training Setup}
            \centering 
            \resizebox{\textwidth}{!}{
            \begin{tabular}[t]{cc}
                \toprule
                Parameter & Value \\
                \midrule
                \# of Epochs & 100 \\
                Batch Size & 128 \\
                Learning Rate & 5e-4 \\
                Layer Decay & 0.65 \\
                Weight Decay & 0.05 \\
                Warmup Epochs & 5 \\
                LR Scheduler & Cosine \\
                Optimizer & AdamW \\
                \bottomrule
            \end{tabular}
            }
        \end{subtable}
    \end{minipage}
\end{table}

\section{QM9 Quantum Property Prediction}\label{app:qm9}

For further comparison, we test \ourmethod{} on the QM9 quantum property prediction task. We use the same training setup used by \citet{stark20223dinfomax}. Note that 3DInfomax previously used the PNA~\cite{corso2020pna} architecture, which performs graph convolution with multiple message aggregators. As reference, we provide a list of baselines previously tested in \cite{stark20223dinfomax} below:
\begin{itemize}
    \item \textsc{RandInit}: Randomly initialized weights.
    \item \textsc{GraphCL}~\cite{you2020graphcl}: Contrastive loss to align representations under graph augmentations.
    \item \textsc{ProPred}: Predicting the Gibbs free energy
    \item \textsc{DisPred}: Predicting all atom-pairwise distances
    \item \textsc{ConfGen}~\cite{ganea2021geomol}: Generating 10 conformers per molecule
\end{itemize}
Note that these baselines are trained on the GEOM-Drugs pretraining subset consisted of 140k molecules unlike 3DInfomax and \ourmethod{}, which are trained on a 50k subset of QM9. For \ourmethod{}, we train a TorchMD-NET on the same pretraining subset as our 3D teacher, and distill its representations to PNA.

Table~\ref{tab:vs_3dinfomax_qm9} shows the full results. We find that similar to \textsc{3DInfomax}, \textsc{DnD-Node} consistently results in positive knowledge transfer, as it outperforms \textsc{RandInit} across all QM9 tasks. When comparing to \textsc{3DInfomax}, we find significant performance boosts in 6 out of 8 tasks, and competitive accuracy in the other two. We find that the largest performance gain occurs in the \textsc{alpha} task. This shows that our approach of distilling representations from 3D conformers to 2D encoders is valid not only for attention-based graph encoding architectures, but for convolution-based GNNs as well.

\begin{table}[!ht]
    \centering
    \caption{Test evaluation results on each QM9 quantum property prediction dataset. The baseline results are from \citealt{stark20223dinfomax}. \ourmethod{} results are from using the same set of seeds used by the baselines. Best results and results within one standard deviation are in \textbf{bold}.} \label{tab:vs_3dinfomax_qm9} 
    \vspace{.1in}
    \resizebox{\textwidth}{!}{\begin{tabular}{l|cccccccc}
        \toprule
        Target & $\mu$ & $\alpha$ & \textsc{homo} & \textsc{lumo} & \textsc{gap} & \textsc{r2} & \textsc{zpve} & $c_v$\\
        Metric & MAE($\downarrow$) & MAE($\downarrow$) & MAE($\downarrow$) & MAE($\downarrow$) & MAE($\downarrow$) & MAE($\downarrow$) & MAE($\downarrow$) & MAE($\downarrow$)\\
        \midrule
        \textsc{RandInit} & 0.4133 & 0.3972 & 82.10 & 85.72 & 123.08 & 22.14 & 15.08 & 0.1670 \\
        \textsc{GraphCL} & 0.3937 & 0.3295 & 79.57 & 80.81 & 120.08 & 21.84 & 12.39 & 0.1422 \\
        \textsc{ProPred} & 0.3975 & 0.3732 & 93.11 & 99.84 & 131.99 & 29.21 & 11.17 & 0.1795 \\
        \textsc{DisPred} & 0.4626 & 0.3570 & 80.58 & 84.93 & 116.21 & 29.23 & 25.91 & 0.1587 \\
        \textsc{ConfGen} & 0.3940 & 0.4219 & 79.75 & 79.16 & 110.72 & 20.86 & 21.10 & 0.1555 \\
        \textsc{3DInfomax} & \bf 0.3507{\scriptsize $\pm$0.005} & 0.3268{\scriptsize $\pm$0.006} & \bf 68.96{\scriptsize $\pm$0.32} & \bf 69.51{\scriptsize $\pm$0.54} & 101.71{\scriptsize $\pm$2.03} & \bf 17.39{\scriptsize $\pm$0.94} & \bf 7.966{\scriptsize $\pm$1.87} & 0.1306{\scriptsize $\pm$0.009} \\
        \midrule
        \textsc{D\&D-Graph} & \bf 0.3512{\scriptsize $\pm$0.005} & 0.2903{\scriptsize $\pm$0.029} & 70.36{\scriptsize $\pm$2.70} & 71.72{\scriptsize $\pm$2.17} & 98.82{\scriptsize $\pm$1.09} & \bf 17.61{\scriptsize $\pm$0.68} & 12.88{\scriptsize $\pm$3.42} & \bf 0.1248{\scriptsize $\pm$0.006}\\
        \textsc{D\&D-Node} & 0.3552{\scriptsize $\pm$0.004} & \bf 0.2807{\scriptsize $\pm$0.015} & \bf 69.32{\scriptsize $\pm$1.01} & \bf 69.63{\scriptsize $\pm$0.62} & \bf 98.79{\scriptsize $\pm$0.89} & \bf 17.75{\scriptsize $\pm$0.43} & 10.19{\scriptsize $\pm$1.77} & 0.1429{\scriptsize $\pm$0.015}\\
        \bottomrule
    \end{tabular}
    }
\end{table}



\end{document}